\documentclass[10pt,twocolumn,letterpaper]{article}

\usepackage[pagenumbers]{cvpr} 

\usepackage{graphicx}
\usepackage{amsmath}
\usepackage{amssymb}
\usepackage{booktabs}
\usepackage{multirow}
\usepackage{color}
\usepackage{colortbl}
\definecolor{best}{RGB}{255, 153, 153}
\definecolor{second}{RGB}{255, 204, 153}
\usepackage{enumitem}
\usepackage{gensymb}
\usepackage[pagebackref,breaklinks,colorlinks]{hyperref}

\usepackage[capitalize]{cleveref}

\usepackage[capitalize]{cleveref}
\crefname{section}{Sec.}{Secs.}
\Crefname{section}{Section}{Sections}
\Crefname{table}{Table}{Tables}
\crefname{table}{Tab.}{Tabs.}

\def\cvprPaperID{5615} 
\def\confName{CVPR}
\def\confYear{2023}

\begin{document}

\title{UV Volumes for Real-time Rendering of Editable Free-view Human Performance}

\renewcommand{\thefootnote}{\fnsymbol{footnote}} 
\author{Yue Chen$^{1}$\footnotemark[1] \qquad Xuan Wang$^{2,3}$\footnotemark[1] \qquad Xingyu Chen$^{1}$ \qquad  Qi Zhang$^3$ \\
\qquad Xiaoyu Li$^{3}$ \qquad Yu Guo$^1$\footnotemark[2] \qquad Jue Wang$^3$ \qquad Fei Wang$^1$ \vspace{3pt}\\
$^{1}$Xi'an Jiaotong University\ \ \ \  $^{2}$Ant Group\ \ \ \  $^{3}$Tencent AI Lab\\
}

\twocolumn[{%
\renewcommand\twocolumn[1][]{#1}%
\maketitle

\vspace{-0.8cm}
\begin{center}
    \includegraphics[width=0.95\hsize]{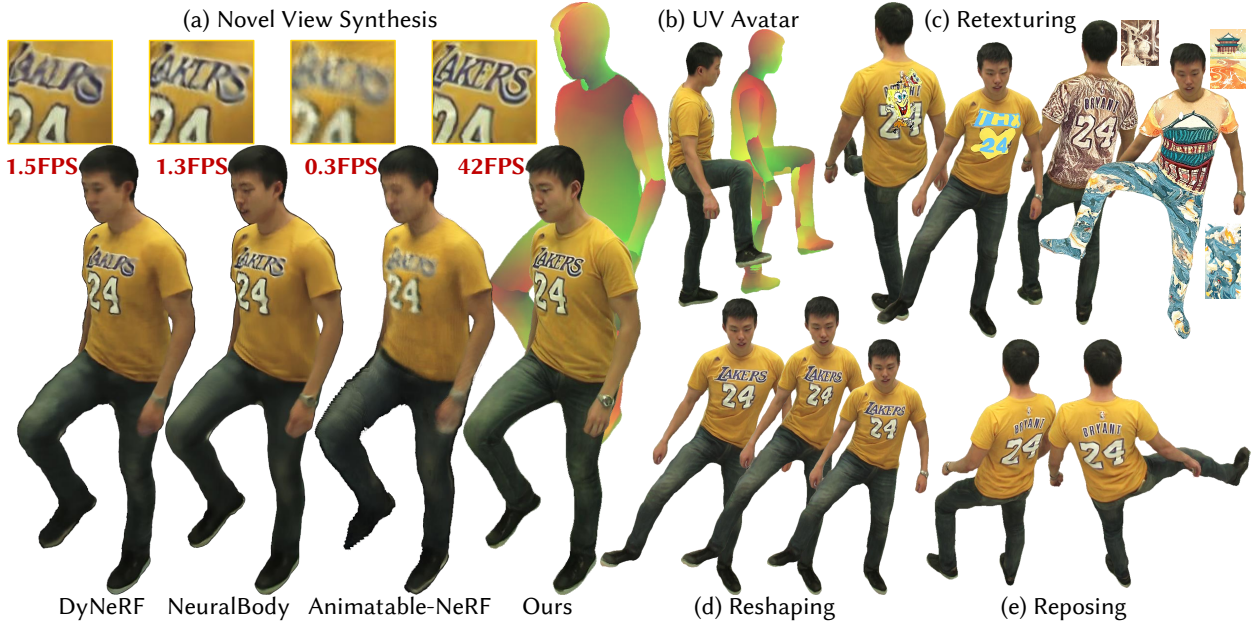}
    \vspace{-5pt}
    \captionof{figure}{We decompose a dynamic human into 3D UV Volumes along with a 2D texture. The disentanglement of appearance and geometry enables us to achieve (a) high-fidelity real-time novel view synthesis guided by (b) a smooth UV avatar, (c) retexturing of a 3D human by editing a 2D texture, (d) reshaping and (e) reposing by changing the parameters of a human model while keeping the texture untouched.}
    \label{fig:teaser}
\end{center}
}]

\footnotetext[1]{Authors contributed equally to this work.}
\footnotetext[2]{Corresponding Author.}

\begin{abstract}
\vspace{-5pt}
Neural volume rendering enables photo-realistic renderings of a human performer in free-view, a critical task in immersive VR/AR applications. 
But the practice is severely limited by high computational costs in the rendering process. 
To solve this problem, we propose the {\em UV Volumes}, a new approach that can render an editable free-view video of a human performer in real-time. 
It separates the high-frequency (i.e., non-smooth) human appearance from the 3D volume, and encodes them into 2D neural texture stacks (NTS). 
The smooth UV volumes allow much smaller and shallower neural networks to obtain densities and texture coordinates in 3D while capturing detailed appearance in 2D NTS.
For editability, the mapping between the parameterized human model and the smooth texture coordinates allows us a better generalization on novel poses and shapes.
Furthermore, the use of NTS enables interesting applications, e.g., retexturing. Extensive experiments on CMU Panoptic, ZJU Mocap, and H36M datasets show that our model can render 960 $\times$ 540 images in 30FPS on average with comparable photo-realism to state-of-the-art methods.
The project and supplementary materials are available at \href{https://fanegg.github.io/UV-Volumes}{https://fanegg.github.io/UV-Volumes}.
\vspace{-8pt}
\end{abstract}

\section{Introduction}
Synthesizing a free-view video of a human performer in motion is a long-standing problem in computer vision. Early approaches~\cite{fvv} rely on obtaining an accurate 3D mesh sequence through multi-view stereo. However, the computed 3D mesh often fails to depict the complex geometry structure, resulting in limited photorealism. In recent years, methods (e.g., NeRF~\cite{mildenhall2020nerf}) that make use of volumetric representation and differentiable ray casting have shown promising results for novel view synthesis. These techniques have been further extended to tackle dynamic scenes.

Nonetheless, NeRF and its variants require a large number of queries against a deep Multi-Layer Perceptron (MLP). Such time-consuming computation prevents them from being applied to applications that require high rendering efficiency. In the case of static NeRF, a few methods~\cite{garbin2021fastnerf,reiser2021kilonerf,yu2021plenoctrees} have already achieved real-time performance. However, for dynamic NeRF, solutions for real-time rendering of volumetric free-view video are still lacking. 

In this work, we present {\em UV Volumes}, a novel framework that can produce an editable free-view video of a human performer in motion and render it in real-time. Specifically, we take advantage of a pre-defined UV-unwrapping (e.g., SMPL or dense pose) of the human body to tackle the geometry (with texture coordinates) and textures in two branches. We employ a sparse 3D Convolutional Neural Networks (CNN) to transform the voxelized and structured latent codes anchored with a posed SMPL model to a 3D feature volume, in which only smooth and view-independent densities and UV coordinates are encoded. For rendering efficiency, we use a shallow MLP to decode the density and integrate the feature into the image plane by volume rendering. Each feature in the image plane is then individually converted to the UV coordinates. Accordingly, we utilize the yielded UV coordinates to query the RGB value from a pose-dependent neural texture stack (NTS). This process greatly reduces the number of queries against MLPs and enables real-time rendering.

\begin{figure}[t!]
\centering
  \includegraphics[width=1.0\linewidth]{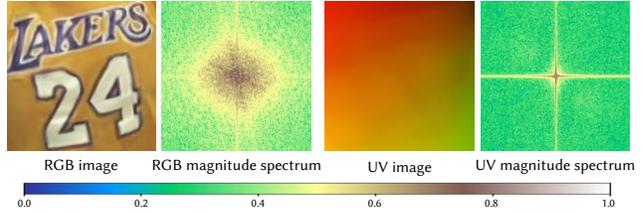}
   \caption{Discrete Fourier Transform (DFT) for RGB and UV image. In the magnitude spectrum, the distance from each point to the midpoint describes the frequency, the direction from each point to the midpoint describes the direction of the plane wave, and the value of the point describes its amplitude. The distribution of the UV magnitude spectrum is more concentrated in the center, which indicates that the frequency of the UV image is lower.
   }
\label{fig:magnitude_spectrum}
\end{figure}

It is worth noting that the 3D Volumes in the proposed framework only need to approximate relatively ``smooth'' signals. As shown in Figure \ref{fig:magnitude_spectrum}, the magnitude spectrum of the RGB image and the corresponding UV image indicates that UV is much smoother than RGB.
That is, we only model the low-frequency density and UV coordinate in the 3D volumes, and then detail the appearance in the 2D NTS, which is also spatially aligned across different poses. 
The disentanglement also enhances the generalization ability of such modules and supports various editing operations. 

We perform extensive experiments on three widely-used datasets: CMU Panoptic, ZJU Mocap, and H36M datasets. The results show that the proposed approach can effectively generate an editable free-view video from both dense and sparse views. The produced free-view video can be rendered in real-time with comparable photorealism to the state-of-the-art methods that have much higher computational costs. In summary, our major contributions are:

\begin{itemize}[leftmargin=24pt]
	\item A novel system for rendering editable human performance video in free-view and real-time.
	\item UV Volumes, a method that can accelerate the rendering process while preserving high-frequency details.
	\item Extended editing applications enabled by this framework, such as reposing, retexturing, and reshaping.
\end{itemize}

\begin{figure*}[t]
\centering
  \includegraphics[width=1\linewidth]{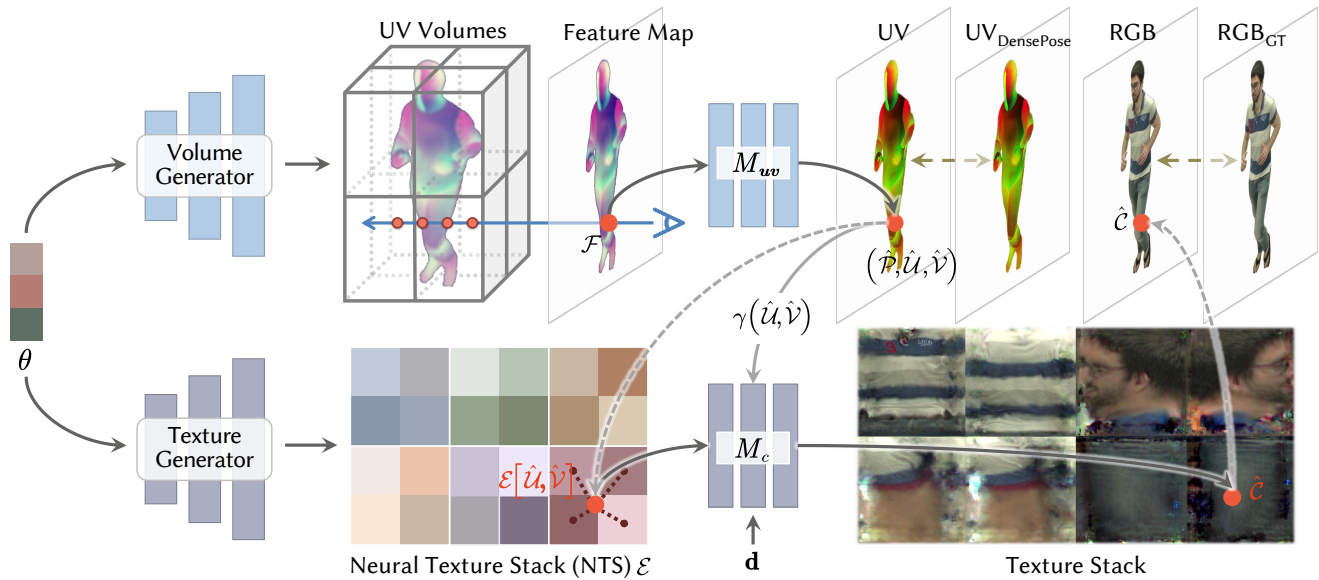}
   \caption{Overall pipeline of proposed framework. Our model has two main branches: 1) Based on a human pose $\theta$, a volume generator constructs UV volumes involving the feature of UV information. Then a feature map can be rendered via differentiable raymarching and decoded to texture coordinates (UV) pixel-by-pixel. 2) A texture generator produces a pose-dependent Neural Texture Stack(NTS) $\mathcal{E}$ that encodes the highly-detailed appearance information. The UV coordinates and the texture embedding interpolated from NTS are passed into an MLP to predict the color $\hat{\mathcal{C}}$ at the desired ray direction $\mathbf{d}$.}
   \vspace{-10pt}
\label{fig:pipe}
\end{figure*}

\section{Related Work}
\paragraph{\textbf{Novel View Synthesis for Static Scenes.}}
Novel view synthesis for static scenes is a well-explored problem. Early image-based rendering approaches~\cite{davis2012unstructured,gortler1996lumigraph,levoy1996light,buehler2001unstructured,eisemann2008floating} utilize densely sampled images to obtain novel views with light fields instead of explicit or accurate geometry estimation. The learning-based methods~\cite{flynn2019deepview,kalantari2016learning,mildenhall2019local,srinivasan2019pushing,hedman2018deep} apply neural networks to reuse input pixels from observed viewpoints. 
In recent years, dramatic improvements have been achieved by neural volume rendering techniques. For instance, NeRF~\cite{mildenhall2020nerf} represents a static scene using a deep MLP, mapping 3D spatial locations and 2D viewing directions to volumetric density and radiance. 
For computation efficiency, rendering high-resolution scenes via NeRF is time-consuming since it requires millions of queries to obtain the density and radiance. Subsequent works~\cite{garbin2021fastnerf,reiser2021kilonerf,yu2021plenoctrees,yu2021plenoxels,hedman2021baking,niemeyer2021giraffe} attempt to accelerate the inference of vanilla NeRF in various ways, some of which achieve real-time rendering performance, but only for static scenes. For editability, the generative models, FENeRF~\cite{fenerf} and IDE-3D~\cite{ide3D}, exploit semantic masks to edit the synthesized free-view portraits, but they are not compatible with the free-view performance capture task. NeuTex~\cite{neutex} also employs the UV-texture to store the appearance and enables editing on the texture map. Unfortunately, it can only tackle static objects.

\paragraph{\textbf{Free-View Video Synthesis.}}
Early methods~\cite{mustafa2016temporally, collet2015high} rely on accurate 3D reconstruction and texture rendering captured by dome-based multi-camera systems to synthesize novel views of a dynamic scene. 
Recently, various neural representations have been employed in differentiable rendering to depict dynamic scenes, such as voxels~\cite{lombardi2019neural}, point clouds~\cite{wu2020multi}, textured meshes~\cite{thies2019deferred,bagautdinov2021driving,ma2021pixel}, and implicit functions~\cite{peng2021neural,liu2020neural,li2021neural,pumarola2021d,park2021nerfies,park2021hypernerf}. 
Particularly, DyNeRF~\cite{li2021neural} takes the latent code as the condition for time-varying scenes, while NeuralBody~\cite{peng2021neural} employs structured latent codes anchored to a posed human model. 
Other deformation-based NeRF variants~\cite{pumarola2021d,park2021nerfies,tretschk2021non,park2021hypernerf,li2021neural2} take as input the monocular video, as a result, they fail to synthesize the free-view spatio-temporal visual effects. Besides, they also suffer from the high computational cost in inference and the lack of editing abilities. 
Geometric constraints and discrete space representation are exploited in methods~\cite{shao2021doublefield,yu2021function4d}, and a hybrid scene representation is used for efficiency in ~\cite{MVP,remelli2022drivable}. The method~\cite{foctree} employs Fourier PlenOctree to accelerate rendering, but the photorealism is harmed by the shared discrete representation across the time sequence. Furthermore, all these models are still non-editable.

\paragraph{\textbf{Editable Free-View Videos.}}
There exist previous works that focus on the problem of producing editable free-view videos or animatable avatars. ST-NeRF~\cite{zhang2021editable} exploits the layered neural representation in order to move, rotate and resize individual objects in free-view videos.
Some methods~\cite{peng2021animatable,weng2022humannerf,xu2021h}
decompose a dynamic human into a canonical neural radiance field and a skeleton-driven warp field that backward maps observation-space points to canonical space. However, learning a backward warp field is highly under-constrained since the backward warp field is pose-dependent~\cite{chen2021snarf}. 
Textured Neural Avatars~\cite{shysheya2019textured} proposes utilizing the texture map to improve novel pose generalization, whereas employing a 2D rendering neural network prevents it from consistent novel view synthesis. Neural Actor~\cite{neuralactor} takes the texture map as latent variables. Nevertheless, the requirement of the ground-truth texture map limits their application in many cases.
In contrast, our approach estimates the texture map end-to-end, and can produce editable (including reposing, reshaping and retexturing) free-view videos in real-time from both dense and sparse views.

\section{Method}

Given multi-view videos of a performer, our model generates an editable free-view video that supports real-time rendering. We use the availability of an off-the-shelf SMPL model and the pre-defined UV unwrap in Densepose \cite{densepose} to introduce proper priors into our framework. In this section, we describe the details of our framework, which is shown in Figure ~\ref{fig:pipe}. The two main branches in our framework are presented in turn. One is to generate the UV volumes (Sec.~\ref{sec:uvvolume}), and the other is the generation of NTS (Sec.~\ref{sec:texture}). Then we provide a more detailed description of the training process in Sec.~\ref{sec:training}.

\subsection{UV Volumes} \label{section:UV}
\label{sec:uvvolume}
Neural radiance fields~\cite{mildenhall2020nerf} have been proven to produce free-viewpoint images with view consistency and high fidelity. Nonetheless, capturing the high-fidelity appearance in a dynamic scene is time-consuming and difficult. To this end, we propose the UV volumes in which only the density and texture coordinate (i.e., UV coordinate) are encoded instead of human appearance. 
Given the UV image rendered by ray casting, we can use the UV coordinates to query the corresponding RGB values from the 2D NTS by employing the UV unwrap defined in Densepose.

We utilize the volume generator to construct UV volumes. First, the time-invariant latent codes anchored to a posed SMPL model are voxelized and taken as the input. Then we use the 3D sparse CNN to encode the voxelized latent codes to a 3D feature volume named UV volumes, which contains UV information. 

Given a sample image $\mathcal{I}$ of multi-view videos, we provide a posed SMPL parameterized by human pose $\theta$ and a set of latent codes $\mathbf{z}$ anchored on its vertices and then query the feature vector $f\!\left(\mathbf{x}, \mathbf{z}, \theta\right)$ at point $\mathbf{x}$ from the generated UV volumes. The feature vector is fed into a shallow MLP $M_{\sigma}$ to predict the volume density:
\begin{equation}
\label{eqn:01}
\sigma(\mathbf{x})=M_{\sigma}\!\left(f\!\left(\mathbf{x}, \mathbf{z}, \theta\right)\right).
\end{equation}

We then apply the volume rendering~\cite{kajiya1984ray} technique to render the UV feature volume into a 2D feature map. 
We sample $N_{i}$ points $\left\{\mathbf{x}_{i}\right\}_{i=1}^{N_{i}}$ along the camera ray $\mathbf{r}$ between near and far bounds based on the posed SMPL model in 3D space. The feature at the pixel can be calculated as:
\begin{equation}
\label{eqn:02}
\begin{gathered}
\mathcal{F}(\mathbf{r})=\sum_{i=1}^{N_{i}} T_{i}\left(1-\exp \!\left(-\sigma\!\left(\mathbf{x}_{i}\right) \delta_{i}\right)\right)f\!\left(\mathbf{x_{i}}, \mathbf{z}, \theta\right), \\
\text{where  } T_{i}=\exp \left(-\sum_{j=1}^{i-1} \sigma\!\left(\mathbf{x}_{j}\right) \delta_{j}\right),
\end{gathered}
\end{equation}
and $\delta_{i}=\left\|\mathbf{x}_{i+1}-\mathbf{x}_{i}\right\|_{2}$ is the distance between adjacent sampled points. An MLP $M_{uv}$ is then used to individually decode all the pixels in the yielded view-invariant feature map to their corresponding texture coordinates and generate the UV image. In specific, the texture coordinates can be represented as:
\begin{equation}
\label{eqn:03}
\left(\hat{\mathcal{P}}(\mathbf{r}),\hat{\mathcal{U}}(\mathbf{r}),\hat{\mathcal{V}}(\mathbf{r})\right) =M_{uv}\!\left(\mathcal{F}(\mathbf{r})\right),
\end{equation}
where $\hat{\mathcal{P}}$ and $\hat{\mathcal{U}},\hat{\mathcal{V}}$ are the corresponding part assignments and UV coordinates, respectively.

\subsection{Neural Texture Stack} \label{section:Texture}
\label{sec:texture}
Given the generated UV image, we employ the continuous texture stack encoded in the implicit neural representation to recover the color image. To extract the local relation of the neural texture stack with respect to the human pose, we use a CNN texture generator $G$ to produce the pose-dependent NTS:
\begin{equation}
\label{eqn:04}
\mathcal{E}_{k}=G\!\left(\theta, \mathbf{k}\right),
\end{equation}
where we subdivide the body surface into $N_k=24$ parts, and $\mathbf{k}$ is a one-hot label vector representing the $k$-th body part. At a foreground pixel, the part assignments $\hat{\mathcal{P}}$ predicted from UV volumes (referred in Equation~\eqref{eqn:03}) can be interpreted as the probability of the pixel belonging to the $k$-th body part, which is defined as $\sum_{k=1}^{N_k} \hat{\mathcal{P}}_{k}(\mathbf{r})=1$. 
For each human body part $k$, the texture generator generates the corresponding neural texture stack $\mathcal{E}_k$. 
We forward propagate the generator network $G$ once to predict the neural textures with a batch size of $24$.
Let $\hat{\mathcal{U}}_{k}$ and $\hat{\mathcal{V}}_{k}$ denote the predicted UV coordinates of the $k$-th body part. 
We sample the texture embeddings at non-integer locations $(\hat{\mathcal{U}}_{k}(\mathbf{r}), \hat{\mathcal{V}}_{k}(\mathbf{r}))$ in a piecewise-differentiable manner using bilinear interpolation~\cite{jaderberg2015spatial}:
\begin{equation}
\label{eqn:05}
\mathbf{e}_k(\mathbf{r})=\mathcal{E}_{k}\!\left[\hat{\mathcal{U}}_{k}(\mathbf{r}),\hat{\mathcal{V}}_{k}(\mathbf{r})\right].
\end{equation}

To model the high-frequency color of human performances, we apply positional encoding $\gamma(\cdot)$~\cite{rahaman2019spectral} to UV coordinates and the viewing direction, and pass the encoded UV map along with the sampled texture embedding into an MLP $M_c$ to decode the view-dependent color $\hat{\mathcal{C}}_{k}(\mathbf{r})$ of camera ray $\mathbf{r}$ at the desired viewing direction $\mathbf{d}$:
\begin{equation}
\label{eqn:07}
\hat{\mathcal{C}}_{k}(\mathbf{r})= {M}_{c}\!\left(\gamma(\hat{\mathcal{U}}_{k}(\mathbf{r}),\hat{\mathcal{V}}_{k}(\mathbf{r})),\mathbf{e}_{k}(\mathbf{r}),\mathbf{k},\gamma(\mathbf{d})\right).
\end{equation}

Following that, the color $\hat{\mathcal{C}}(\mathbf{r})$ at each pixel is reconstructed via a weighted combination of decoded colors at $N_{k}$ body parts, where the weights are prescribed by part assignments $\hat{\mathcal{P}}_{k}$:
\begin{equation}
\hat{\mathcal{C}}(\mathbf{r})=\sum_{k=1}^{N_{k}}\hat{\mathcal{P}}_{k}(\mathbf{r}) \, \hat{\mathcal{C}}_{k}(\mathbf{r}).
\end{equation}

\begin{figure}[tb]
\centering
  \includegraphics[width=1\linewidth]{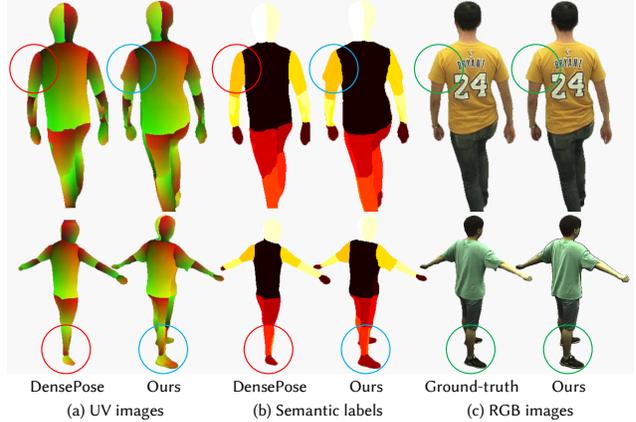}
   \caption{Given noisy UV and semantic labels (e.g., red circles), we can recover proper UV volumes (e.g., blue circles) under the intrinsic multi-view constraint of minimizing the photometric error between renderings and ground-truth (e.g., green circles).}
\label{fig:iuv_gt}
\end{figure}

\begin{figure*}[t!]
\centering
  \includegraphics[width=1.0\linewidth]{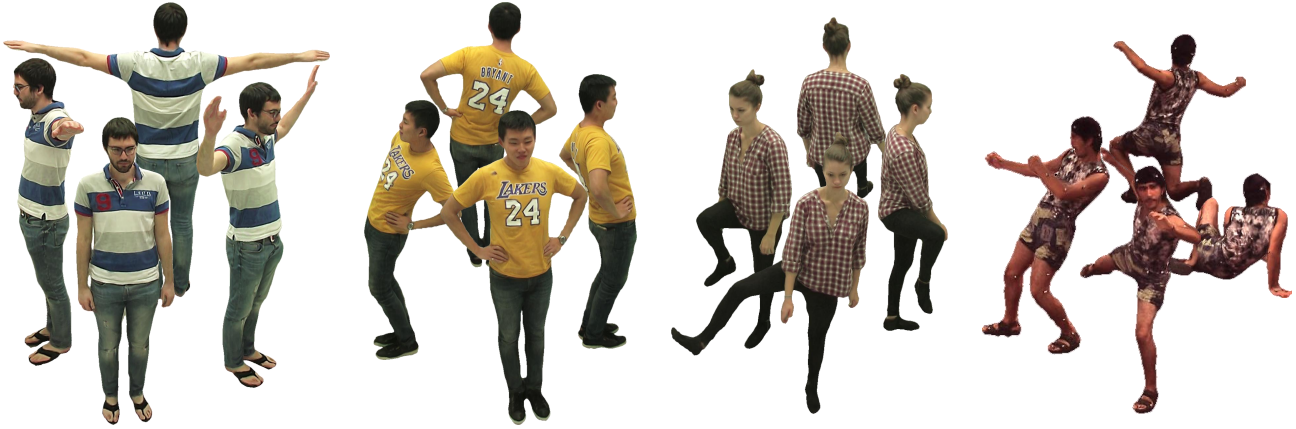}
   \caption{The novel view synthesis of our model on various human performances, which achieves high-fidelity renderings in real-time.
   }
  \vspace{-10pt}
\label{fig:view_pose}
\end{figure*}

\subsection{Training}
\label{sec:training}
Collecting the results of all rays $\{\hat{\mathcal{C}}(\mathbf{r})\}^{H \times W}$, we denote the entire rendered image as $\hat{\mathcal{I}} \in \mathbb{R}^{H \times W \times 3}$. To learn the parameters of our model, we optimize the photometric error between the renderings $\hat{\mathcal{I}}$ and the ground-truth images $\mathcal{I}$:
\begin{equation}
\mathcal{L}_{\text{rgb}}=\left\|\hat{\mathcal{I}}-\mathcal{I}\right\|_{2}^{2}.
\end{equation}

Benefiting from our memory-saving framework that disentangles appearance and geometry, we can render an entire image during training instead of sampling image patches~\cite{mildenhall2020nerf, peng2021neural}.
Thus, we also compare the rendered images against the ground-truth using perceptual loss~\cite{gatys2016image, johnson2016perceptual, ulyanov2016texture}, which extracts feature maps by a pretrained fixed VGG network $\psi(\cdot)$~\cite{simonyan2014very} from both images and minimizes the L1-norm between them:
\begin{equation}
\label{eqn:9}
\mathcal{L}_{\text{vgg}}=\left\|\psi(\hat{\mathcal{I}})-\psi(\mathcal{I})\right\|_{1}.
\end{equation}

To warm-start the UV volumes and regularize its solution space, we leverage the pre-trained DensePose model as an auxiliary supervisor. In particular, we perform the DensePose network on the training data and utilize the outputs of Denspose as pseudo supervision, such that we can regularize UV volumes by semantic loss $\mathcal{L}_{\text{p}}$ and UV-metric loss $\mathcal{L}_{\text{uv}}$ between DensePose outputs and our UV images:
\begin{equation}
\begin{aligned}
\label{eqn:10}
\mathcal{L}_{\text{p}}&=\sum_{k=1}^{N_k} \mathcal{P}_{k} \log (\hat{\mathcal{P}}_{k}), \\
\mathcal{L}_{\text{uv}}&=\sum_{k=1}^{N_k} \mathcal{P}_{k} 
\left(
\left\|\hat{\mathcal{U}}_{k}-\mathcal{U}_{k}\right\|_{2}^{2}
+\left\|\hat{\mathcal{V}}_{k}-\mathcal{V}_{k}\right\|_{2}^{2}
\right),
\end{aligned}
\end{equation}
where $N_k$ is the number of body parts, and $\mathcal{P}_{k}$ and $\hat{\mathcal{P}}_{k}$ are respectively the multi-class semantic probability at the $k$-th part of DensePose outputs and UV images. Similarly, $\mathcal{U}_{k}$,$\mathcal{V}_{k}$ and $\hat{\mathcal{U}}_{k}$, $\hat{\mathcal{V}}_{k}$ are the predicted UV coordinates at the $k$-th part of DensePose and UV images, respectively.
$\mathcal{L}_{\text{p}}$ is chosen as a multi-class cross-entropy loss to encourage rendered part labels to be consistent with provided DensePose labels, and $\mathcal{L}_{\text{uv}}$ promotes to generate inter-frame consistent UV coordinates.

We present the UV images predicted by our UV volumes and the pseudo supervision of DensePose in Figure \ref{fig:iuv_gt}. Given noisy semantic and UV labels (e.g., the red circles), we can reconstruct proper UV volumes (e.g., the blue circles) under the intrinsic multi-view constraint of RGB loss (e.g., the green circles). As shown in the second row of Figure \ref{fig:iuv_gt}, it can be observed that UV volumes successfully recover the UV images even though the provided DensePose supervision is incorrect.

Given the binary human mask $\mathcal{S}$ for the observed image $\mathcal{I}$, we propose a silhouette loss to facilitate UV volumes modeling a more fine-grained geometry:
\begin{equation}
\begin{aligned}
\label{eqn:11}
T(\mathbf{r})&=\exp \left(-\!\sum_{j=1}^{N_{i}-1} \sigma(\mathbf{x}_{j}) \delta_{j}\right),
\\
\mathcal{L}_{\text{s}}&= \sum_{\mathbf{r} \in \mathcal{R}} \left(\mathcal{S}(\mathbf{r})(1-T(\mathbf{r}))+(1-\mathcal{S}(\mathbf{r}))T(\mathbf{r})\right), 
\end{aligned}
\end{equation}
where $T(\mathbf{r})$ is accumulated transmittance. Here we define the value of mask $\mathcal{S}(\mathbf{r})$ in the foreground as zero, and the background as one.

We combine the aforementioned losses and jointly train our model to optimize the full objective:
\begin{equation}
\label{eqn:12}
\mathcal{L}=\mathcal{L}_{\text{rgb}}+\lambda_{\text{vgg}}\mathcal{L}_{\text{vgg}}+\lambda_{\text{p}}\mathcal{L}_{\text{p}}+\lambda_{\text{uv}}\mathcal{L}_{\text{uv}}+\lambda_{\text{s}}\mathcal{L}_{\text{s}}.
\end{equation}

\begin{table*}[t!]
    \centering
    \setlength\tabcolsep{3pt}
    \resizebox{\linewidth}{!}{
        \begin{tabular}{c|c||ccccc|ccccc|ccccc||cccc}
            \toprule
            \multicolumn{2}{c||}{\multirow{3}{*}{Datasets}} & \multicolumn{15}{c||}{View synthesis quality} & \multicolumn{4}{c}{Efficiency} \vspace{1.5pt} \\
            \multicolumn{2}{c||}{} & \multicolumn{5}{c|}{PSNR $\uparrow$} & \multicolumn{5}{c|}{SSIM $\uparrow$} & \multicolumn{5}{c||}{LPIPS $\downarrow$} & \multicolumn{4}{c}{FPS $\uparrow$} \\
            \cmidrule{3-21}
            \multicolumn{2}{c||}{} 
            & \small {DN}
            & \small {NB} 
            & \small {AN}
            & \footnotesize {w/o $\mathcal{L}_{p}$}
            & \small {Ours}
            & \small {DN}
            & \small {NB} 
            & \small {AN}
            & \footnotesize {w/o $\mathcal{L}_{p}$}
            & \small {Ours}
            & \small {DN}
            & \small {NB} 
            & \small {AN}
            & \footnotesize {w/o $\mathcal{L}_{p}$}
            & \small {Ours}
            & \small {DN}
            & \small {NB} 
            & \small {AN}
            & \small {Ours} \\
            \midrule
            \multicolumn{1}{c|}{\multirow{3}{*}{\begin{tabular}[c]{@{}c@{}}CMU\\ (960$\times$540)\end{tabular}}} 
            & p1 & 30.04 & 29.78 & 27.12 & 30.09 & \cellcolor{second}{30.38} & \cellcolor{second}{0.968} & 0.962 & 0.936 & 0.963 & 0.966 & 0.088 & 0.099 & 0.135 & 0.055 & \cellcolor{second}{0.036} & 1.01 & 0.76 & 0.21 & \cellcolor{second}{44.76} \\
            & p2 & 25.56 & 25.68 & 26.13 & 28.51 & \cellcolor{second}{28.78} & 0.939 & 0.942 & 0.903 & 0.952 & \cellcolor{second}{0.953} & 0.137 & 0.139 & 0.204 & 0.062 & \cellcolor{second}{0.044} & 1.45 & 1.28 & 0.34 & \cellcolor{second}{37.30} \\
            & p3 & 27.04 & 27.12 & 24.20 & 29.36 & \cellcolor{second}{29.38} & 0.955 & 0.956 & 0.874 & \cellcolor{second}{0.962} & \cellcolor{second}{0.962} & 0.154 & 0.142 & 0.259 & 0.062 & \cellcolor{second}{0.047} & 2.12 & 1.28 & 0.33 & \cellcolor{second}{34.60} \\
            \midrule
            \multicolumn{1}{c|}{\multirow{3}{*}{\begin{tabular}[c]{@{}c@{}}ZJU\\ (512$\times$512)\end{tabular}}} & 313 & \cellcolor{second}{29.67} & 28.82 & 27.50 & 28.44 & 29.11 & \cellcolor{second}{0.958} & 0.952 & 0.939 & 0.956 & \cellcolor{second}{0.958} & 0.084 & 0.088 & 0.124 & 0.068 & \cellcolor{second}{0.053} & 2.07 & 1.51 & 0.62 & \cellcolor{second}{51.39} \\
            & 377 & 27.13 & \cellcolor{second}{28.12} & 25.71 & 26.18 & 26.28 & 0.933 & \cellcolor{second}{0.949} & 0.923 & 0.931 & 0.930 & 0.112 & 0.088 & 0.152 & 0.094 & \cellcolor{second}{0.085} & 2.41 & 2.02 & 0.76 & \cellcolor{second}{38.70} \\
            & 386 & \cellcolor{second}{30.29} & 30.12 & 28.51 & 28.38 & 28.48 & 0.938 & \cellcolor{second}{0.939} & 0.915 & 0.919 & 0.916 & 0.122 & 0.112 & 0.163 & 0.103 & \cellcolor{second}{0.078} & 3.00 & 4.89 & 0.91 & \cellcolor{second}{35.88} \\
            \midrule
            \multicolumn{1}{c|}{\multirow{3}{*}{\begin{tabular}[c]{@{}c@{}}H36M\\ (500$\times$500)\end{tabular}}} & s9p & 21.53 & 25.11 & 26.08 & 26.03 & \cellcolor{second}{26.19} & 0.824 & 0.912 & \cellcolor{second}{0.917} & 0.915 & 0.916 & 0.242 & 0.136 & 0.139 & 0.085 & \cellcolor{second}{0.084} & 1.06 & 2.19 & 0.30 & \cellcolor{second}{40.00} \\
            & s11p & 21.27 & 24.39 & 25.21 & 25.20 & \cellcolor{second}{25.82} & 0.828 & 0.899          & 0.906 & 0.905 & \cellcolor{second}{0.911} & 0.313 & 0.193 & 0.174 & 0.118 & \cellcolor{second}{0.111} & 1.18 & 1.02 & 0.67 & \cellcolor{second}{33.41} \\
            & s1p & 18.91 & 23.24 & 23.43 & 23.83 & \cellcolor{second}{23.98} & 0.781 & 0.909 & 0.901 & \cellcolor{second}{0.911} & \cellcolor{second}{0.911} & 0.332 & 0.149 & 0.162 & 0.094 & \cellcolor{second}{0.093} & 1.38 & 0.97 & 0.50 & \cellcolor{second}{41.43} \\
            \bottomrule 
        \end{tabular}
    }
    \vspace{-4pt}
    \caption{
        Quantitative results of \textbf{novel view synthesis}. We present competitive PSNR and SSIM while outperforming baselines on LPIPS (agrees well with human visual perception~\cite{zhang2018perceptual}) and achieve 30 FPS (pre-computed sparse CNN) available for real-time applications.
    }
    \vspace{-4pt}
    \label{tab:novel_view_tab}
\end{table*}

\begin{table*}[t!]
    \centering
    \begin{minipage}{0.535\linewidth}
        \includegraphics[width=1\linewidth,page=1]{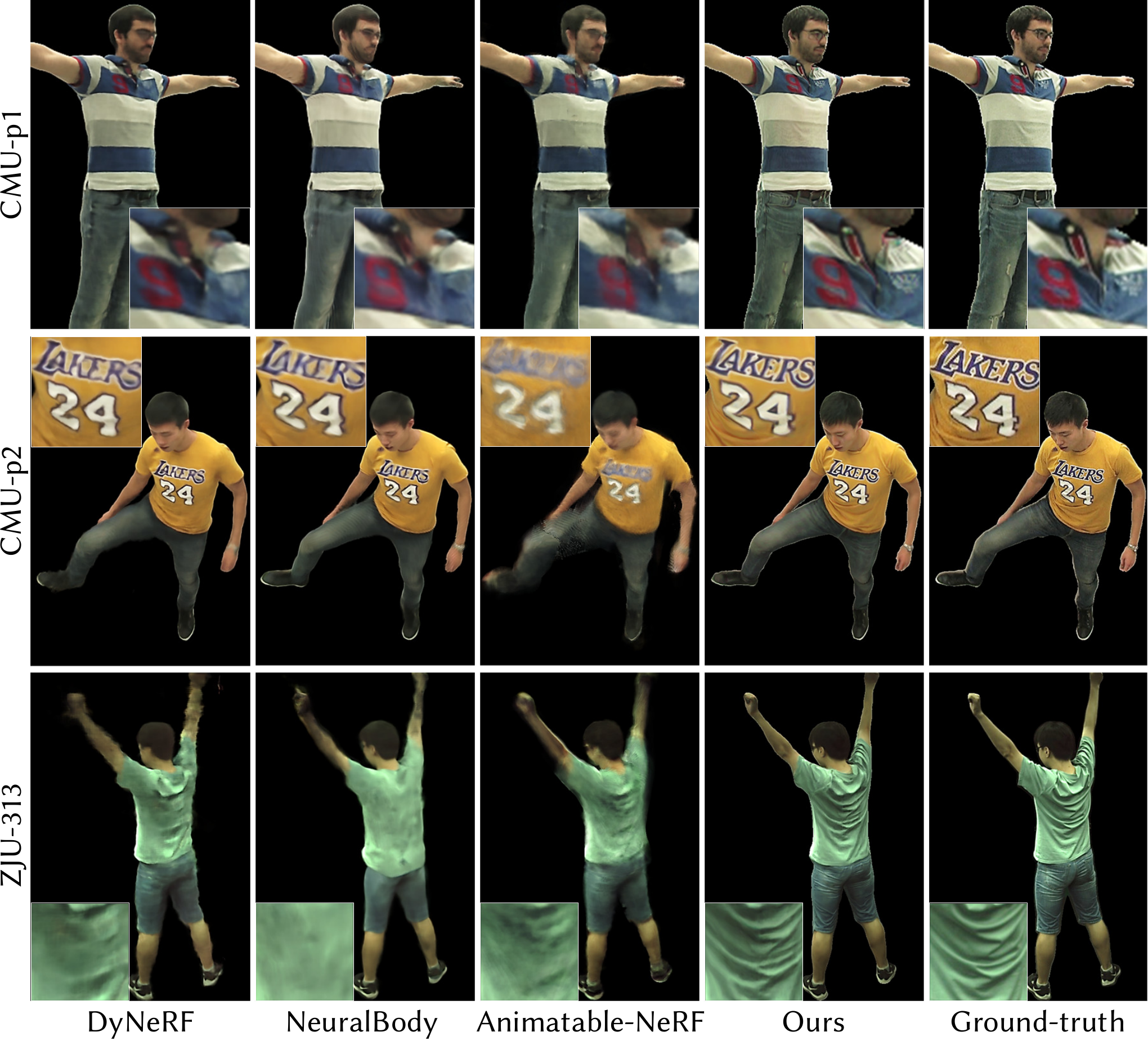}
        \vspace{-16pt}
        \captionof{figure}{
            Qualitative results of \textbf{novel view synthesis} on CMU Panoptic and ZJU Mocap. 
            Benefiting from spatially aligning the appearance across different poses in a 2D texture, our method produces high-fidelity novel view synthesis, while baselines suffer from blurs (at letters and wrinkles).
        }
        \vspace{-4pt}
        \label{fig:novel_view}
    \end{minipage}
    \hspace{8pt}
    \begin{minipage}{0.435\linewidth}
        \includegraphics[width=1\linewidth,page=1]{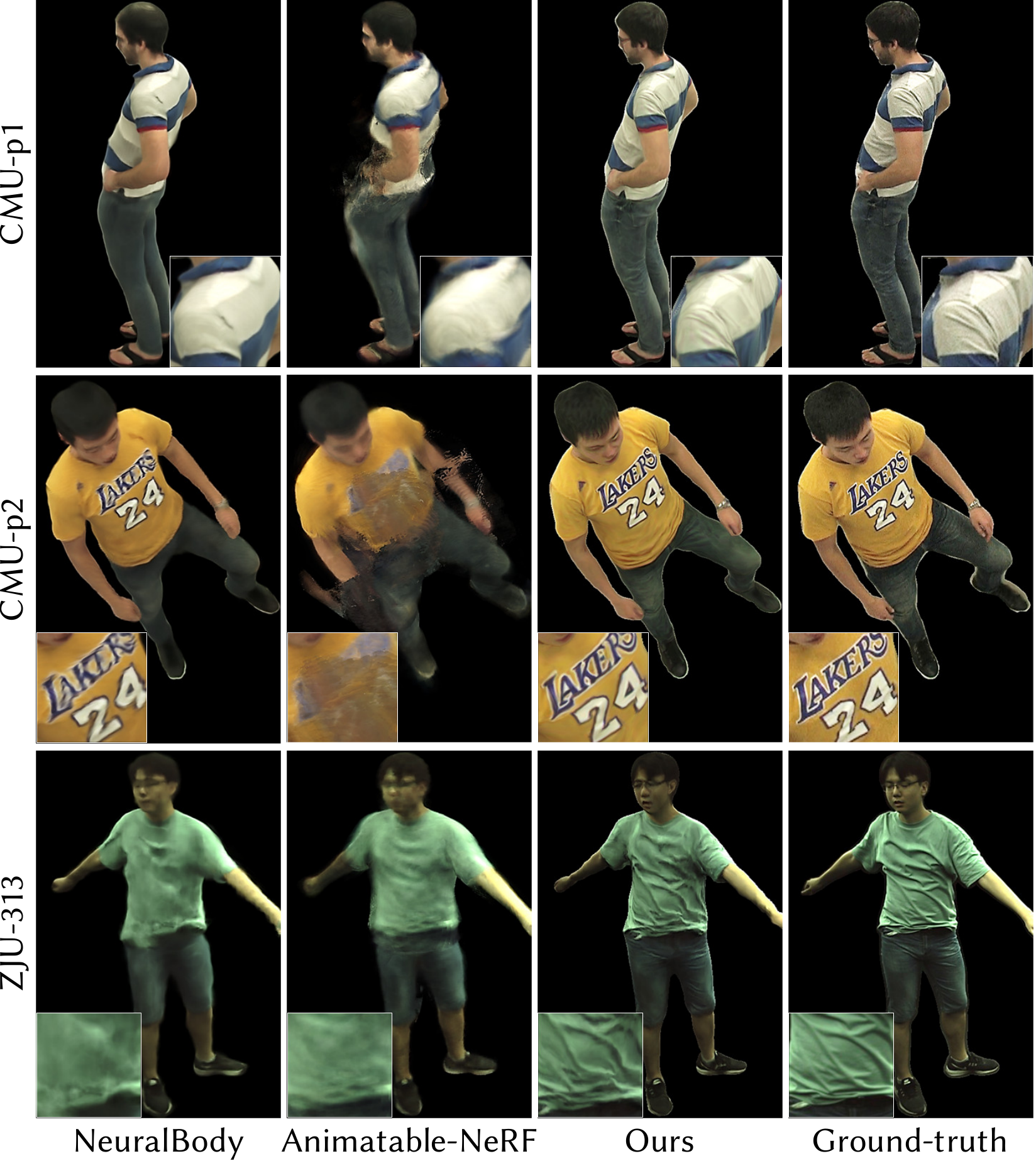}
        \vspace{-16pt}
        \captionof{figure}{
                Qualitative results of \textbf{novel pose synthesis} on CMU Panoptic, ZJU Mocap. 
                Benefiting from the disentanglement of appearance and geometry, our method performs better on novel poses, especially for preserving sharp details.
        }
        \label{fig:novel_pose}
        \vspace{-4pt}
    \end{minipage}
\end{table*}

\renewcommand{\thefootnote}{\arabic{footnote}}

\section{Experiments}
\label{sec:exp}
To demonstrate the effectiveness and efficiency of our method, we perform extensive experiments. We report quantitative results using four standard metrics: PSNR, SSIM, LPIPS, and FPS\footnote{The sparse CNN output is pre-computed for the reported framerates.}. And the qualitative experiments further illustrate that our method produces photo-realistic images in different tasks, e.g., novel view synthesis, reposing, reshaping, and retexturing.

\noindent\textbf{Dataset.}
We perform experiments on several types of datasets which consist of calibrated and synchronized multi-view videos. We use 26 and 20 training views on CMU Panoptic dataset~\cite{CMUDATA} with $960\times540$ resolution and ZJU Mocap dataset~\cite{peng2021neural} with $512\times512$ resolution, respectively. The most challenging one is the H36M dataset~\cite{h36m} with $500\times500$ resolution, where only three cameras are available for training. We obtain the binary human mask by ~\cite{gong2018instance}. The evaluation is done on the hold-out cameras (novel views) or hold-out segments of the sequence (novel poses).

\noindent\textbf{Baselines.}
To validate our method, we compare it against several state-of-the-art free-view video synthesis techniques: 1) DN: DyNeRF~\cite{li2021neural}, which takes time-varying latent codes as the conditions for dynamic scenes; and 2) NB: NeuralBody~\cite{peng2021neural}, which takes as input the posed human model with structured time-invariant latent codes and generates a pose-conditioned neural radiance field; 3) AN: Animatable-NeRF~\cite{peng2021animatable}, which uses neural blend weight fields to generate correspondences between observation and canonical space.

\begin{table}[t]
\centering
\setlength\tabcolsep{3pt}
\resizebox{\linewidth}{!}{
    \begin{tabular}{c||ccc|ccc|ccc}
        \toprule
        \multirow{2}{*}{Method} & \multicolumn{3}{c|}{CMU (960$\times$540)} & \multicolumn{3}{c|}{ZJU (512$\times$512)} & \multicolumn{3}{c}{H36M (500$\times$500)} \vspace{1.5pt} \\
        & \small {PSNR $\uparrow$} & \small {SSIM $\uparrow$} & \small {LPIPS $\downarrow$} & \small {PSNR $\uparrow$} & \small {SSIM $\uparrow$} & \small {LPIPS $\downarrow$} & \small {PSNR $\uparrow$} & \small {SSIM $\uparrow$} & \small {LPIPS $\downarrow$} \\
        \midrule
        NB & 25.94 & 0.918 & 0.146 & 24.51 & \cellcolor{second}{0.918} & 0.120 & \cellcolor{second}{25.54} & \cellcolor{second}{0.884} & 0.170 \\
        AN & 23.65 & 0.883 & 0.208 & \cellcolor{second}{24.55} & 0.911 & 0.153 & 25.00 & 0.873 & 0.170 \\
        Ours & \cellcolor{second}{26.20} & \cellcolor{second}{0.927} & \cellcolor{second}{0.073} & 23.69 & 0.910 & \cellcolor{second}{0.104} & 25.04 & 0.874 & \cellcolor{second}{0.141} \\
        \bottomrule 
    \end{tabular}
}
\vspace{-4pt}
\captionof{table}{
    Quantitative results of \textbf{novel pose synthesis}. We achieve competitive PSNR and SSIM while outperforming baselines on LPIPS, which agrees well with humans~\cite{zhang2018perceptual}.
}
\vspace{-8pt}
\label{tab:novel_pose_tab}
\end{table}

\noindent\textbf{Novel View Synthesis.}
For comparison, we synthesize images of training poses in hold-out test views. Table \ref{tab:novel_view_tab} shows the comparison of our method against baselines, which demonstrates that our method performs best LPIPS and FPS among all methods. Specifically, we achieve rendering free-view videos of human performances in 30FPS with the help of UV volumes. 
Note that LPIPS agrees surprisingly well with human visual perception~\cite{zhang2018perceptual}, which indicates that our synthesis is more visually similar to ground-truth.

Figure \ref{fig:novel_view} presents the qualitative comparison of our method with baselines. Baselines fail to preserve the sharp image details, whose rendering is blurry and even split. In contrast, our method can accurately capture high-frequency details like letters, numbers and wrinkles on shirts and the belt on pants benefiting from our NTS model. 
Furthermore, we show the view synthesis results of dynamic humans in Figure \ref{fig:view_pose},
which indicate that our method generates high-quality appearance results even with rich textures and challenging motions. Note that the rightmost example is from the H36M dataset with only 4 views. Please refer to the supplementary material for more results.

\noindent\textbf{Reposing.}
We perform reposing on the human performer with novel motions. As DyNeRF is not designed for editing tasks, we compare our method against NeuralBody and Animatable-NeRF. As shown in Table \ref{tab:novel_pose_tab}, quantitative results demonstrate that our method achieves competitive PSNR and SSIM while outperforming others on LPIPS.

The qualitative results are shown in Figure \ref{fig:novel_pose}. For novel human poses, NeuralBody gives blurry and distorted rendering results, while Animatable-NeRF even produces split humans due to a highly under-constrained backward warp field from observation to canonical space. In contrast, synthesized images of our method exhibit better visual quality with reasonable high-definition dynamic textures. The results indicate that using smooth UV volumes in 3D and encoding texture in 2D has better controllability on the novel pose generalization than directly modeling a pose-conditioned neural radiance field.

\begin{figure}[t!]
\centering
  \includegraphics[width=1\linewidth]{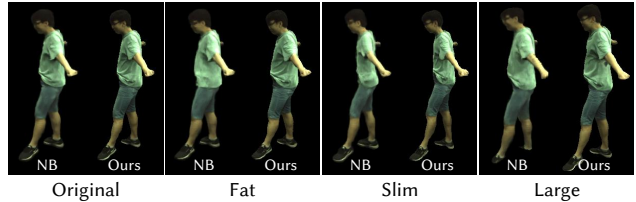}
    \vspace{-16pt}
  \caption{Qualitative results of \textbf{reshaping}. By changing the SMPL parameters $\beta$, we can conveniently make the human performer fatter, slimmer, or larger. The result of NeuralBody is shown on the left of each image pair, while ours is on the right. Obviously, more details and consistency are preserved by ours in varying shapes.}
\label{fig:edit_shape}
\end{figure}

\noindent\textbf{Reshaping.} 
We demonstrate that our approach can edit the shape of reconstructed human performance by changing the shape parameters of the SMPL model. We illustrate the qualitative results in Figure \ref{fig:teaser} and Figure \ref{fig:edit_shape}. NeuralBody fails to infer the reasonable changes of the cloth, while our method generalizes well on novel shapes.

\begin{figure}[t!]
\centering
  \includegraphics[width=1\linewidth]{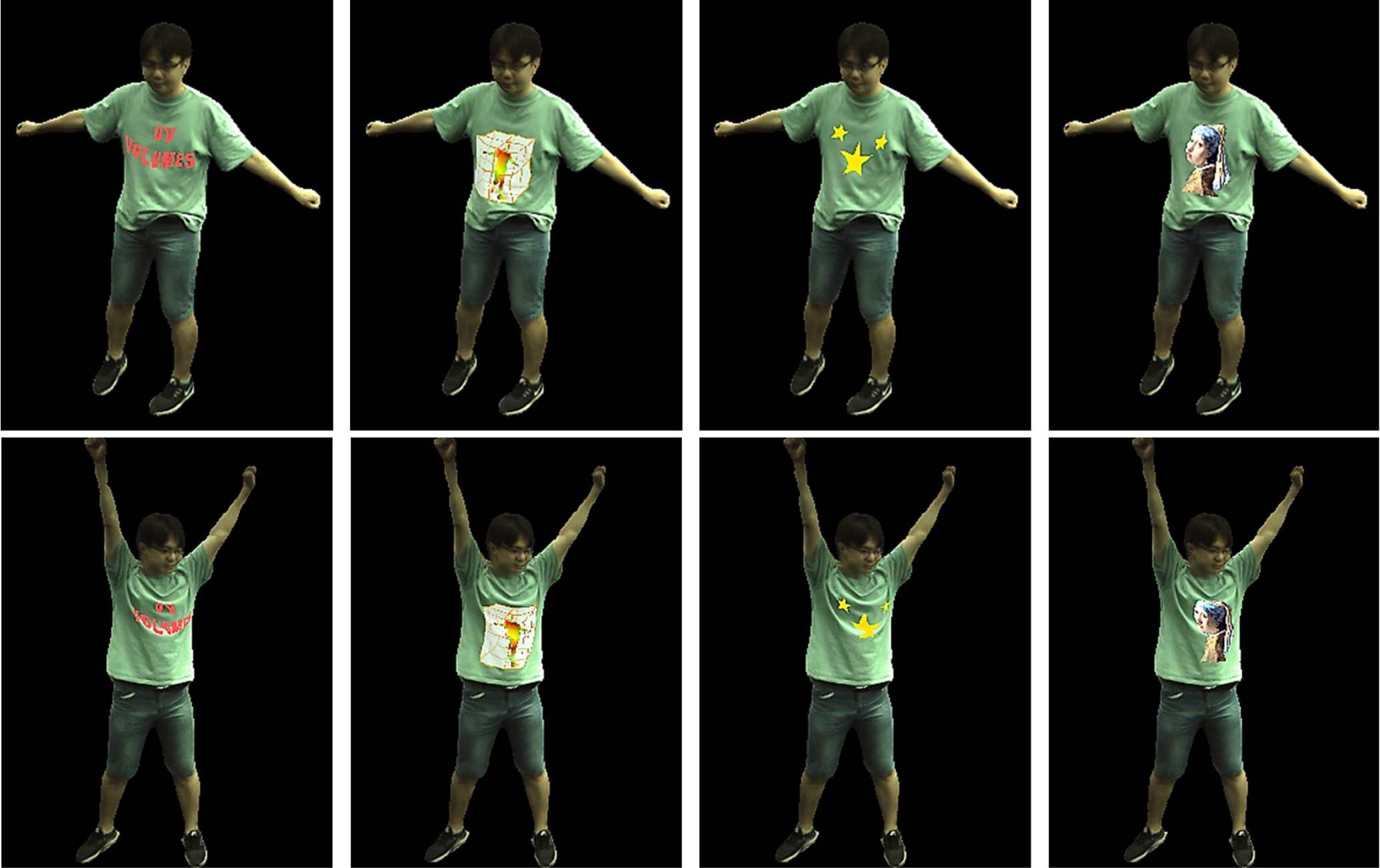}
  \vspace{-16pt}
  \caption{Qualitative results of \textbf{retexturing}. 
  The disentanglement of appearance and geometry
  allows us to conveniently edit the texture by drawing patterns on the NTS. The rich texture patterns are well preserved and transferred to correct semantic areas in different poses, which demonstrates that 
  the texture is not only changed as expected under the edited frame,
  but also transferred to a novel frame with the modeled dynamics.
  }
  \vspace{-6pt}
\label{fig:edit_tex}
\end{figure}

\begin{figure}[t!]
\centering
  \includegraphics[width=1\linewidth]{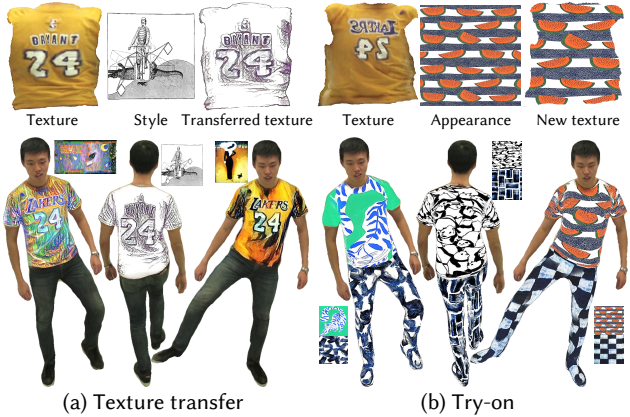}
  \vspace{-12pt}
  \caption{Given any arbitrary artistic style or cloth appearance, we can render (a) a 3D dynamic human with the transferred texture or perform (b) a 3D virtual try-on in real-time.}
\label{fig:transfer}
\vspace{-2pt}
\end{figure}

\noindent\textbf{Retexturing.}
With the learned dense correspondence of UV volumes and neural texture, we can edit the 3D cloth with a user-provided 2D texture, as shown in Figure \ref{fig:edit_tex}. Visually inspected, the rich texture patterns are well preserved and transferred to correct semantic areas in different poses. Moreover, our model supports changing textures' style and appearance, which are presented in Figure \ref{fig:transfer}. Thanks to the style transfer network~\cite{ghiasi2017exploring}, we can perform arbitrary artistic stylizations on 3D human performance. Given any fabric texture, we can even dress the performer in various appearances, which enables 3D virtual try-on in real-time.

\subsection{Ablation Studies}
We conduct ablation studies on performer p1 of the CMU dataset. As shown in Table \ref{tab:Ablation}, we analyze the effects of different losses for the proposed approach by removing warm-start loss, perceptual loss and silhouette loss, respectively.
Then, we analyze the time consumption of each module. We encourage the reader to see the supplement for additional ablations, discussion of model design, and other experimental results.

\begin{table}[tb]
\centering
\setlength\tabcolsep{3pt}
\resizebox{\linewidth}{!}{
    \begin{tabular}{c||ccc|ccc}
        \toprule
        \multirow{2}{*}{Ablations} & \multicolumn{3}{c|}{Novel View Synthesis} & \multicolumn{3}{c}{Novel Pose Generation} \vspace{1.5pt} \\
        & \small {PSNR $\uparrow$} & \small {SSIM $\uparrow$} & \small {LPIPS $\downarrow$} & \small {PSNR $\uparrow$} & \small {SSIM $\uparrow$} & \small {LPIPS $\downarrow$} \\
        \midrule
        No Warm-start Loss & 30.37 & 0.964 & 0.060 & 26.14 & 0.917 & 0.076 \\
        No Perceptual Loss & 30.09 & 0.963 & 0.055 & 26.05 & 0.919 & 0.079 \\
        No Silhouette Loss & 17.47 & 0.874 & 0.207 & 16.95 & 0.860 & 0.218 \\
        Complete Model & \cellcolor{second}{30.38} & \cellcolor{second}{0.966} & \cellcolor{second}{0.036} & \cellcolor{second}{26.20} & \cellcolor{second}{0.927} & \cellcolor{second}{0.073} \\
        \bottomrule 
    \end{tabular}
}
\vspace{-4pt}
\captionof{table}{
    Ablation study about different objective functions.
}
\vspace{-8pt}
\label{tab:Ablation}
\end{table}

\noindent\textbf{Impact of warm-start loss.} We present using semantic and UV-metric loss to warm-start the UV volumes and constrain its solution space. To prove the effectiveness of this process, we train an ablation (No Warm-start Loss) built upon our full model by eliminating the warm-start loss.
It gives a lower performance in all metrics, especially the LPIPS increased a lot when rendering novel views.
This comparison indicates that the warm-start loss yields better information reuse of different frames by transforming the observation XYZ coordinates to canonical UV coordinates defined by the consistent semantic and UV-metric loss.

\noindent\textbf{Impact of perceptual loss.} 
In contrast to sampling image patches as baselines, we can render an entire image during training, allowing us to use perceptual loss. 
Table \ref{tab:Ablation} shows that using the same model but training without the perceptual loss (No Perceptual Loss) gives a lower performance in all metrics, especially the PSNR and LPIPS. 
It demonstrates that the perceptual loss is of critical importance to improving the visual quality of synthesized images, which is also reflected in Table~\ref{tab:novel_view_tab} (w/o $\mathcal{L}_{p}$).

\noindent\textbf{Impact of silhouette loss.} To facilitate the UV volumes modeling a more fine-grained geometry, we employ a silhouette loss by using the 2D binary mask of the human performer. We present an ablation (No Silhouette Loss) built upon our full model by eliminating the silhouette loss, as shown in Table \ref{tab:Ablation}. It is obvious that No Silhouette Loss gives the worst performance in all metrics among all ablations. This comparison shows that our geometry does benefit from the silhouette loss, which can be seen in the supplement to get an intuitive visual impression.

\subsection{Time Consumption}
We analyze the time consumption of each module in our framework and the corresponding module in NeuralBody~\cite{peng2021neural} on ZJU Mocap performer 313, as shown in Table \ref{tab:time_cost}. 
On average, it takes 48.78 ms for us to obtain the UV volumes from the posed human model.
Then, our method takes only 19.46 ms (51FPS) to access the free-view renderings, which benefits from the smooth UV volumes that allow using much smaller and shallower MLP to obtain densities and texture coordinates in 3D while capturing detailed appearance in 2D NTS.
On the contrary, NeuralBody spends 663.84 ms (1.5FPS) to synthesize novel views, which prevents it from being used in applications that require running in real-time. Even on the novel pose generalization task, our method can reach 68.23 ms per frame (14FPS) as well. All experiments are run on a single NVIDIA A100 GPU.

\begin{table}[tb]
\centering
\resizebox{\linewidth}{!}{
\begin{tabular}{c||c|c|c|c|c|c}
\toprule
\multirow{3}{*}{Method} & \multicolumn{6}{c}{\small{Novel Pose Generation}} \\ 
\cline{2-7} 
& \multirow{2}{*}{\begin{tabular}[c]{@{}c@{}}\small{Sparse CNN}\end{tabular}} & \multicolumn{5}{c}{\small{Novel View Synthesis}} \\ 
\cline{3-7} 
& & \small{Density} & \multicolumn{3}{c|}{\small{Color Model}} & \small{Rendering} \\
\midrule
\multirow{5}{*}{Ours} & \multirow{4}{*}{48.78} & \multirow{3}{*}{7.08} & \small{UV} & \small{NTS} & \small{RGB} & \multirow{3}{*}{1.73} \\
\cline{4-6}
& & & 1.53 & 7.52 & 1.60 & \\
\cline{4-6}
& & & \multicolumn{3}{c|}{9.12} & \\
\cline{3-7}
& & \multicolumn{5}{c}{19.46} \\
\cline{2-7} 
& \multicolumn{6}{c}{68.23} \\
\midrule
\multirow{3}{*}{NB} & \multirow{2}{*}{52.04} & 84.38 & \multicolumn{3}{c|}{546.81} & 32.65 \\
\cline{3-7}
& & \multicolumn{5}{c}{663.84} \\
\cline{2-7} 
& \multicolumn{6}{c}{715.88} \\
\bottomrule 
\end{tabular}
}
\vspace{-4pt}
\caption{
Time consumption of each module in milliseconds(ms).}
\label{tab:time_cost}
\vspace{-8pt}
\end{table}

\subsection{Limitation}
Our method leverages the SMPL model as a scaffold and DensePose as supervision. Consequently, our method can handle clothing types that roughly fit the human body, but fails to correct the prediction from DensePose when handling long hair, loose clothing, accessories, and photorealistic hands. Therefore, the future work is to utilize explicit cloth models and extra hand tracking. While we use time-invariant structured latent codes to encourage temporally consistent UV, a little perturbation caused by the volume generator may occur in the dynamic human (e.g., some unnatural sliding on the trouser when retexturing the performer). It might be improved by adding temporal consistency loss. Replacing the volume representation with other sparse structures for efficiency is also promising.

\section{Conclusions}
We present the UV volumes for free-view video synthesis of a human performer. It is the first method to generate a real-time free-view video with editing ability. The key is to employ the smooth UV volumes and highly-detailed textures in an implicit neural texture stack. Extensive experiments demonstrate both the effectiveness and efficiency of our method. In addition to improving efficiency, our approach can also support editing, e.g., reposing, reshaping, or retexturing the human performer in the free-view videos.

\renewcommand{\thesection}{\Alph{section}}
\setcounter{section}{0}
Here we provide more implementation details and experimental results. We encourage the reader to view the video results included in the supplementary materials for an intuitive experience of editable free-view human performance.

\section{Network Architectures} 
\subsection{Volume Generator}
\begin{figure}[tb]
\centering
  \includegraphics[width=1\linewidth]{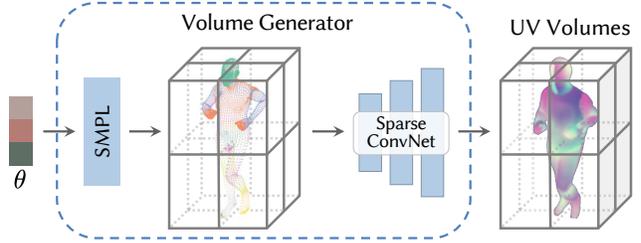}
  \vspace{-20pt}
   \caption{Architecture of the volume generator. It takes human pose $\theta$ as input, drives the point clouds parameterized by the SMPL model under the control of $\theta$, and generates the UV volumes using a 3D sparse CNN to encode a set of latent codes $\mathbf{z}$ anchored on the posed point clouds.}
    \vspace{-5pt}
\label{fig:volume_generator}
\end{figure}

We utilize the volume generator to construct UV volumes, which is presented in Figure \ref{fig:volume_generator}. We take the human pose as input to the SMPL model and animate human point clouds in different poses. Then we follow the previous work \cite{peng2021neural} to anchor a set of time-invariant latent codes to the posed human point cloud and voxelize the point cloud. We follow the network architecture of \cite{peng2021neural} to model the 3D sparse CNN, and reduce the channels from 352 to 64, since the UV volumes only capture low-frequency semantic information. 

\subsection{Density, IUV and Color Network}
\begin{figure}[tb]
\centering
  \includegraphics[width=0.55\linewidth]{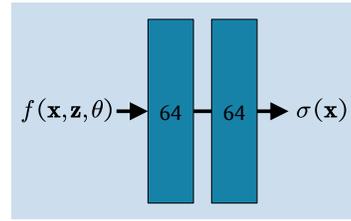}
  \vspace{-8pt}
   \caption{Architecture of the density network. The network takes the feature vector $f\!\left(\mathbf{x}, \mathbf{z}, \theta\right)$ at point $\mathbf{x}$ interpolated by the generated UV volumes and outputs density $\sigma(\mathbf{x})$ using ReLU activation.
   The shallow density MLP $M_{\sigma}$ consists of 2 fully-connected layers with 64 channels.
   }
  \vspace{-5pt}
\label{fig:density_network}
\end{figure}

\begin{figure}[tb]
\centering
  \includegraphics[width=0.9\linewidth]{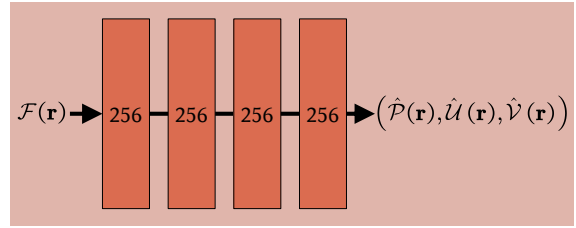}
  \vspace{-8pt}
   \caption{Architecture of the IUV network. 
   The network takes the rendered UV feature vector $\mathcal{F}(\mathbf{r})$ at camera ray $\mathbf{r}$ and outputs view-invariant texture coordinates $\left(\hat{\mathcal{P}}(\mathbf{r}),\hat{\mathcal{U}}(\mathbf{r}),\hat{\mathcal{V}}(\mathbf{r})\right)$ using sigmoid activation. The IUV MLP $M_{uv}$ is modeled by 4 fully-connected layers of 256 channels.}
  \vspace{-5pt}
\label{fig:iuv_network}
\end{figure}

\begin{figure}[t]
\centering
  \includegraphics[width=1\linewidth]{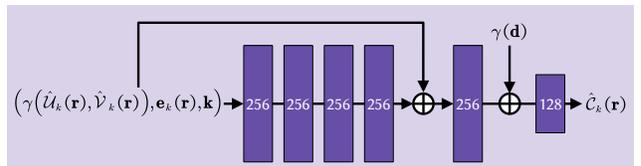}
  \vspace{-20pt}
   \caption{Architecture of the color network. 
   The network takes positional encoding of texture coordinates $\gamma(\hat{\mathcal{U}}_{k}(\mathbf{r}),\hat{\mathcal{V}}_{k}(\mathbf{r}))$ along with the sampled texture embeddings at locations $(\hat{\mathcal{U}}_{k}\!(\mathbf{r}), \hat{\mathcal{V}}_{k}\!(\mathbf{r}))$ and a one-hot part label vector $\mathbf{k}$. The color MLP $M_c$ is modeled by 5 fully-connected layers of 256 channels, including a skip connection that concatenates inputs to the fourth layer’s activation. 
   The feature vector of the fifth layer is processed by an additional layer with 128 channels, along with positional encoding of input viewing direction $\gamma(\mathbf{d})$. A final layer with a sigmoid activation outputs view-dependent RGB color $\hat{\mathcal{C}}_{k}(\mathbf{r})$ of body part $\mathbf{k}$.
   }
  \vspace{-20pt}
\label{fig:rgb_network}
\end{figure}

We present architectures of density network $M_{\sigma}$, IUV network $M_{uv}$ and color network $M_c$ in Figure \ref{fig:density_network}, Figure \ref{fig:iuv_network} and Figure \ref{fig:rgb_network}, respectively. 

\subsection{Texture Generator}
\begin{figure*}[t]
\centering
  \includegraphics[width=1\linewidth]{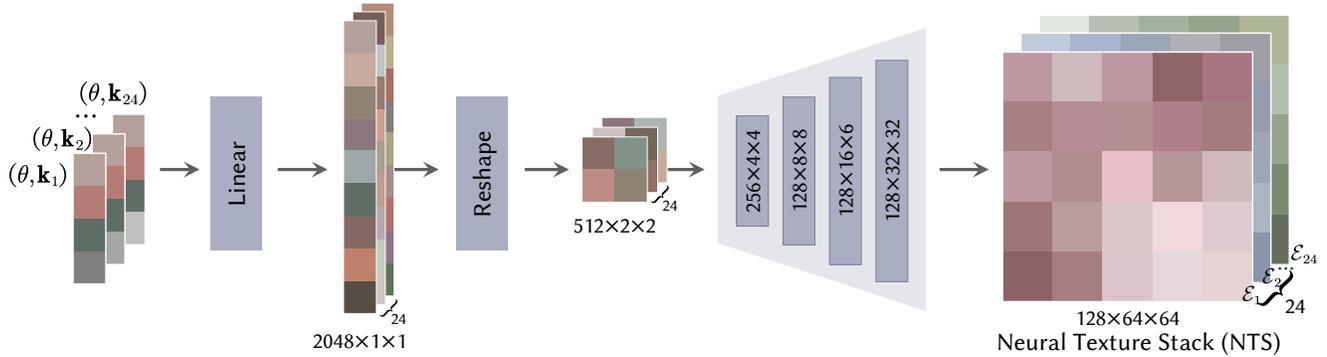}
  \caption{Convolutional texture generator network $G$ consists of 5 convolution layers to get the neural texture stack $\mathcal{E}_i$ with 128 dimensions.}
\label{fig:conv_texture}
\end{figure*}

Figure \ref{fig:conv_texture} shows the architecture of convolutional texture generator network $G$. 
For each human body part of $\{i\}_{i=1}^{24}$, the texture generator generates a corresponding neural texture stack of $\{\mathcal{E}_i\}_{i=1}^{24}$. To predict the specific pose-dependent $\mathcal{E}_i$, we concatenate human pose vector $\theta$ with a one-hot body part label vector $\mathbf{k}_i$ as input to the texture generator.
We forward propagate the generator network $G$ once to predict all the $24$ neural textures with a batch size of $24$.
The CNN-based module is developed to extract the local relation of neural texture stack with respect to the human pose. 
The output spatial neural texture stacks (NTS) will be used for UV unwrapping subsequently.

\section{Additional Implementation Details} 
We set $\lambda_{\text{vgg}}$ to $5\times10^{-2}$ and $\lambda_{\text{s}}$ to $1\times10^{-1}$. The $\lambda_{\text{p}}$ and $\lambda_{\text{uv}}$ are exponential annealing from $1\times10^{-1}$ and $1\times10^{-0}$ to $1\times10^{-3}$ and $5\times10^{-2}$ with $k=4\times10^{-2}$:
\begin{equation}
\begin{aligned}
\lambda_{\text{p}}&=\max(1 \times 10^{-1} e^{-k\cdot \text{epoch}}, 1 \times 10^{-3}) \\
\lambda_{\text{uv}}&=\max(1 \times 10^{-0} e^{-k\cdot \text{epoch}}, 5 \times 10^{-2})
\end{aligned}
\end{equation}

\begin{figure}[t]
    \begin{subfigure}[b]{0.49\linewidth}
        \centering
        \includegraphics[height=2.9cm]{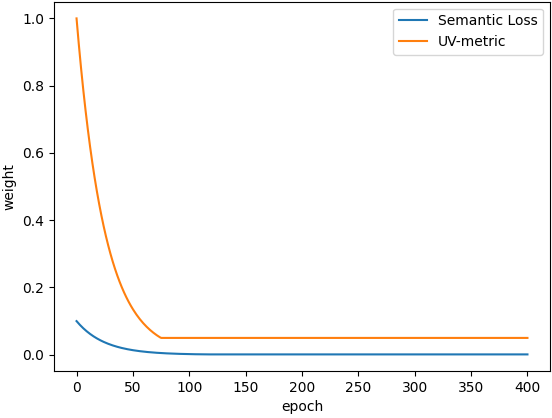}
        \subcaption{Weight of warm-start loss.}
    \label{fig:weight}
    \end{subfigure}
    \begin{subfigure}[b]{0.49\linewidth}
        \centering
        \includegraphics[height=2.9cm]{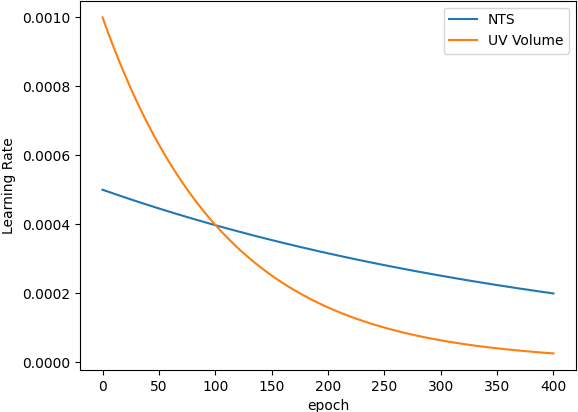}
        \subcaption{Learning rate.}
    \label{fig:lr}
    \end{subfigure}
   \caption{(a) The blue represents the weight of semantic loss $\lambda_{\text{p}}$ and the orange represents the weight of UV-metric loss $\mathcal{L}_{\text{uv}}$. (b) The orange represents the learning rate of the UV volumes branch and the blue represents the learning rate of the NTS branch.}
\end{figure}

As shown in Figure \ref{fig:weight}, the weight of UV-metric is large at the beginning because UV volumes require a warm-start to satisfy the UV unwrap defined by DensePose\cite{guler2018densepose}, and then drop rapidly within 100 epochs because DensePose outputs are not accurate. After 100 epochs, UV-metric becomes a regular term used to constrain the solution space of UV volumes.

UV volumes should be learned primarily in the early stages of training, because The NTS makes sense only after the UV volumes warm-start and a coarse geometry is constructed. Conversely, later in training, we optimize NTS to fit high-frequency signals rather than UV coordinates. Therefore, we use two optimization strategies to train the UV volumes and NTS branch. Their learning rates start from $1\times10^{-3}$ and $5\times10^{-4}$ with a decay rate of $\mathbf{\gamma}=1\times 10^{-1}$, respectively, and decay exponentially along the optimization, as shown in Figure \ref{fig:lr}. 
\begin{equation}
\label{eqn:lr}
\begin{aligned}
l_{\text{nts}}&= 5\times 10^{-4} \mathbf{\gamma}^{\frac{\text{epoch}}{1000}}\\
l_{\text{uv}}&= 1\times 10^{-3} \mathbf{\gamma}^{\frac{\text{epoch}}{250}}
\end{aligned}
\end{equation}

In our experiments, we sample camera rays all over an entire image and 64 points along each ray between near and far bounds. The scene bounds are estimated based on the SMPL model.
We adopt the Adam optimizer\cite{adam} for training our model. We conduct the training on a single NVIDIA A100 GPU. The training on 26-view videos of 100 frames at $960 \times 540$ resolution typically takes around 200k iterations to converge (about 20 hours).

\section{Additional Baseline Method Details} 
\noindent\textbf{DyNeRF (DN)~\cite{li2021neural}.} We reimplement DN by following the procedure in their paper to train the model on video sequences of a moving human.

\noindent\textbf{Neural Body (NB)~\cite{peng2021neural}.} We use the NB code open-sourced by the authors at \href{https://github.com/zju3dv/neuralbody}{https://github.com/zju3dv/neuralbody} and follow their procedure for training on video sequences of a moving human.

\noindent\textbf{Animatable NeRF (AN)~\cite{peng2021animatable}.} We use the AN code open-sourced by the authors at \href{https://github.com/zju3dv/animatable_nerf}{https://github.com/zju3dv/animatable} and follow their procedure for training on video sequences of a moving human.

\section{Additional Ablation Studies}
We conduct ablation studies on performer p1 of the CMU dataset. As shown in Table \ref{tab:ab_tex}, Table \ref{tab:view}, Table \ref{tab:view_frames}, Figure  \ref{fig:ablation_rgb}, Figure \ref{fig:ablation_texture}, and Figure \ref{fig:ab_res}, we analyze the effects of different losses for our proposed approach, different types of NTS, different resolutions of NTS, different methods to model the view-dependent color, and different experimental settings of video frames and input views.

\subsection{Effects of Different Losses}
\noindent\textbf{Impact of warm-start loss.} No Warm-start Loss (w/o $\mathcal{L}_{\text{uv}}$) is an ablation built upon our full model by eliminating the warm-start loss. As shown in Figure  \ref{fig:ablation_rgb} and Figure \ref{fig:ablation_texture}, w/o $\mathcal{L}_{\text{uv}}$ suffers from ambiguity, like the belt on the pants and meaningless texture. This comparison indicates that the warm-start loss yields better information reuse of different frames by transforming the observation XYZ coordinates to canonical UV coordinates defined by the consistent semantic and UV-metric loss.

\noindent\textbf{Impact of perceptual loss.} No Perceptual Loss (w/o $\mathcal{L}_{\text{vgg}}$) is an ablation that uses the same model but training without the perceptual loss. As shown in Figure  \ref{fig:ablation_rgb} and Figure \ref{fig:ablation_texture}, w/o $\mathcal{L}_{\text{vgg}}$ suffers from blur, like the number $9$ on the shirt and distorted number $9$ on the texture. This comparison illustrates that perceptual loss can improve the visual quality of synthesized images by supervising the structure of the renderings from local to global during training.

\noindent\textbf{Impact of silhouette loss.} No Silhouette Loss (w/o $\mathcal{L}_{\text{s}}$) is an ablation built upon our full model by eliminating the silhouette loss. As shown in Figure  \ref{fig:ablation_rgb}, w/o $\mathcal{L}_{\text{s}}$ suffers from artifacts around the performance because there is no the warm-start supervision of semantic and UV-metric labels around the boundary. This comparison demonstrates that silhouette loss is essential for us to model fine-grained geometry.

\subsection{Neural Texture Stacks}
We performed two ablations: 1) different types of NTS; 2) NTS at different resolutions to illuminate the design decisions for the proposed \textit{Neural Texture Stacks}.

\begin{figure}[t]
\centering
  \includegraphics[width=1\linewidth]{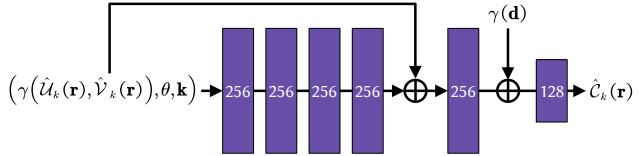}
   \caption{The Global-NTS model directly takes a pose $\theta$ as a condition to generate the color. Its architecture is similar to our color MLP $M_c$ (in Figure \ref{fig:rgb_network}) except for replaces texture embedding $\mathbf{e}_k(\mathbf{r})$ with pose $\theta$.}
\label{fig:global}
\end{figure}

\begin{figure}[t]
\centering
  \includegraphics[width=1\linewidth]{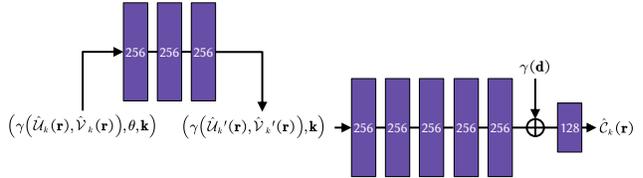}
   \caption{The Local-NTS model transforms UV coordinates $(\hat{\mathcal{U}}_{t}^{k}(\mathbf{r}),\hat{\mathcal{V}}_{t}^{k}(\mathbf{r}))$ to $(\hat{\mathcal{U}}_{t}^{k} \prime(\mathbf{r}), \hat{\mathcal{V}}_{t}^{k} \prime(\mathbf{r}))$ using a deformation field conditioned on pose $\theta$ and modeled by three fully-connected layers of 256 channels. Then we use the transformed UV coordinates, part label vector $\mathbf{k}$ and viewing direction $\mathbf{d}$ as inputs to the subsequent MLP modeled by five fully-connected layers of 256 channels and one fully-connected layer of 128.}
\label{fig:local}
\end{figure}

\begin{figure}[t]
\centering
  \includegraphics[width=1\linewidth]{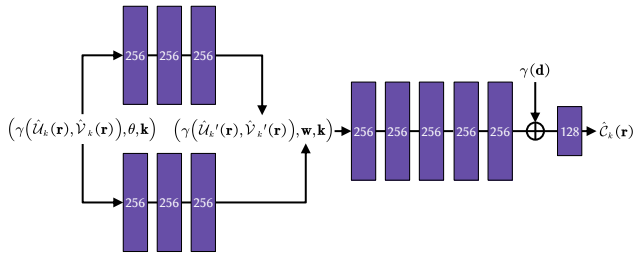}
   \caption{The Hyper-NTS model adds an ambient MLP to the local-NTS model to parameterize the deformation  field, which yields warped UV coordinates $(\hat{\mathcal{U}}_{t}^{k} \prime(\mathbf{r}), \hat{\mathcal{V}}_{t}^{k} \prime(\mathbf{r}))$ and a coordinate $\mathbf{w}$ in the ambient space. Both outputs, part label vector $\mathbf{k}$ and viewing direction $\mathbf{d}$, are concatenated to the subsequent MLP to produce view-dependent colors. The ambient MLP has the same architecture as the deformation MLP.}
\label{fig:hyper}
\end{figure}

\noindent\textbf{Different Types of NTS.} We evaluate our proposed CNN-based Spatial NTS against three ablations: Global NTS, Local NTS, and Hyper NTS. Global NTS (in Figure \ref{fig:global}) is built upon our full model by replacing local texture embedding $\mathbf{e}_k(\mathbf{r})$ with global pose $\theta$. Local NTS (in Figure \ref{fig:local}) transforms observation UV coordinates $(\hat{\mathcal{U}}_{t}^{k}(\mathbf{r}),\hat{\mathcal{V}}_{t}^{k}(\mathbf{r}))$ to canonical UV coordinates $(\hat{\mathcal{U}}_{t}^{k} \prime(\mathbf{r}), \hat{\mathcal{V}}_{t}^{k} \prime(\mathbf{r}))$ using a deformation field. Hyper NTS (in Figure \ref{fig:hyper}) adds an ambient MLP to the local-NTS model to model a slicing surface in hyperspace, which yields a coordinate $\mathbf{w}$ in an ambient space.

\begin{figure*}[!h]
\centering
  \includegraphics[width=1\linewidth]{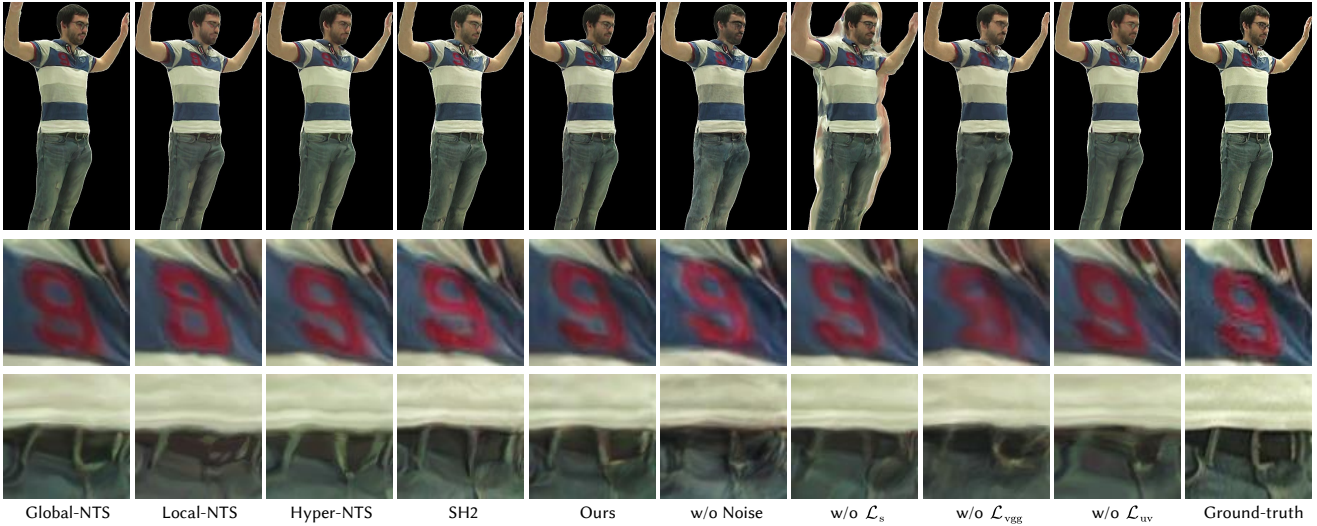}
  \caption{Renderings of our model against ablations.}
\label{fig:ablation_rgb}
\end{figure*}

Table \ref{tab:ab_tex} shows the quantitative results on different types of NTS (i.e., global, local, hyper and spatial). It can be seen that the local-NTS model has the worst performance, which is the most limited among these methods. Local NTS only allows coordinate transformation but cannot generate new topological space, which is totally incapable of modeling the topologically varying texture given different poses. As shown in Figure \ref{fig:ablation_rgb}, it fails to reconstruct the belt on the pants due to the topological variation ((the belt appears when the performer raises his hand and disappears when he puts his hand down because the shirt covers it).

Since the texture of Global-NTS directly conditions on global pose without restriction, it is easy for Global-NTS to generate a new topological space, as shown in Figure \ref{fig:ablation_rgb}, Global-NTS successfully reconstructs the belt on the pants. However, the method that globally models the texture variation is hard to reuse the information of different observation spaces, which leads to ambiguous textures. As shown in Figure \ref{fig:ablation_rgb}, the outline of the number $9$ on the shirt is not clear and even connected, making it look like an $8$. The ambiguous textures are shown in Figure \ref{fig:ablation_texture}, where the number looks more like a red stain.
Lack of local mapping makes a relatively poor performance of Global-NTS as demonstrated in Table \ref{tab:ab_tex}.

Hyper-NTS can model local texture changes by coordinate transformation and generate new topological spaces simultaneously, so it performs better than Global-NTS and Local-NTS. However, it is a thorny issue to tune the dimension of coordinate $\mathbf{w}$. If the dimension of the coordinates $\mathbf{w}$ is too high, Hyper-NTS works as Global-NTS,  while functioning as Local-NTS if too low.

In contrast, as shown in Table \ref{tab:ab_tex}, our CNN-based Spatial NTS outperforms all other NTS, which benefits from the nature of convolution operation capturing local 2D texture changes. At the same time, the MLP only needs to model the local mapping between the neural texture stack and RGB color.

\begin{table}[t]
\centering
\setlength\tabcolsep{3pt}
\resizebox{\linewidth}{!}{
\begin{tabular}{c||ccc|ccc}
\toprule
\multirow{2}{*}{Type} & \multicolumn{3}{c|}{Novel View Synthesis} & \multicolumn{3}{c}{Novel Pose Generation} \vspace{1.5pt} \\
& \small PSNR$\uparrow$ 
& \small SSIM$\uparrow$ 
& \small LPIPS$\downarrow$ 
& \small PSNR$\uparrow$ 
& \small SSIM$\uparrow$ 
& \small LPIPS$\downarrow$ \\
\midrule
Global NTS & 29.86 & 0.966 & 0.062 & 26.18 & \cellcolor{second}{0.927} & 0.074 \\
Local NTS & 29.82 & 0.965 & 0.061 & 26.15 & 0.926 & 0.074 \\
Hyper NTS & 30.08 & \cellcolor{second}{0.966} & 0.059 & 26.19 & \cellcolor{second}{0.927} & \cellcolor{second}{0.073} \\
Spatial NTS & \cellcolor{second}{30.38} & \cellcolor{second}{0.966} & \cellcolor{second}{0.036} & \cellcolor{second}{26.20} & \cellcolor{second}{0.927} & \cellcolor{second}{0.073} \\
\bottomrule 
\end{tabular}
}
\caption{Effects of different types of NTS on our model.}
\vspace{-5pt}
\label{tab:ab_tex}
\end{table}

As demonstrated in Figure \ref{fig:ablation_rgb} and Figure \ref{fig:ablation_texture}, our CNN-based Spatial NTS can accurately capture high-frequency details like numbers or wrinkles on shirts, glasses, and a belt on pants. 

\begin{figure*}[t]
\centering
  \includegraphics[width=1\linewidth]{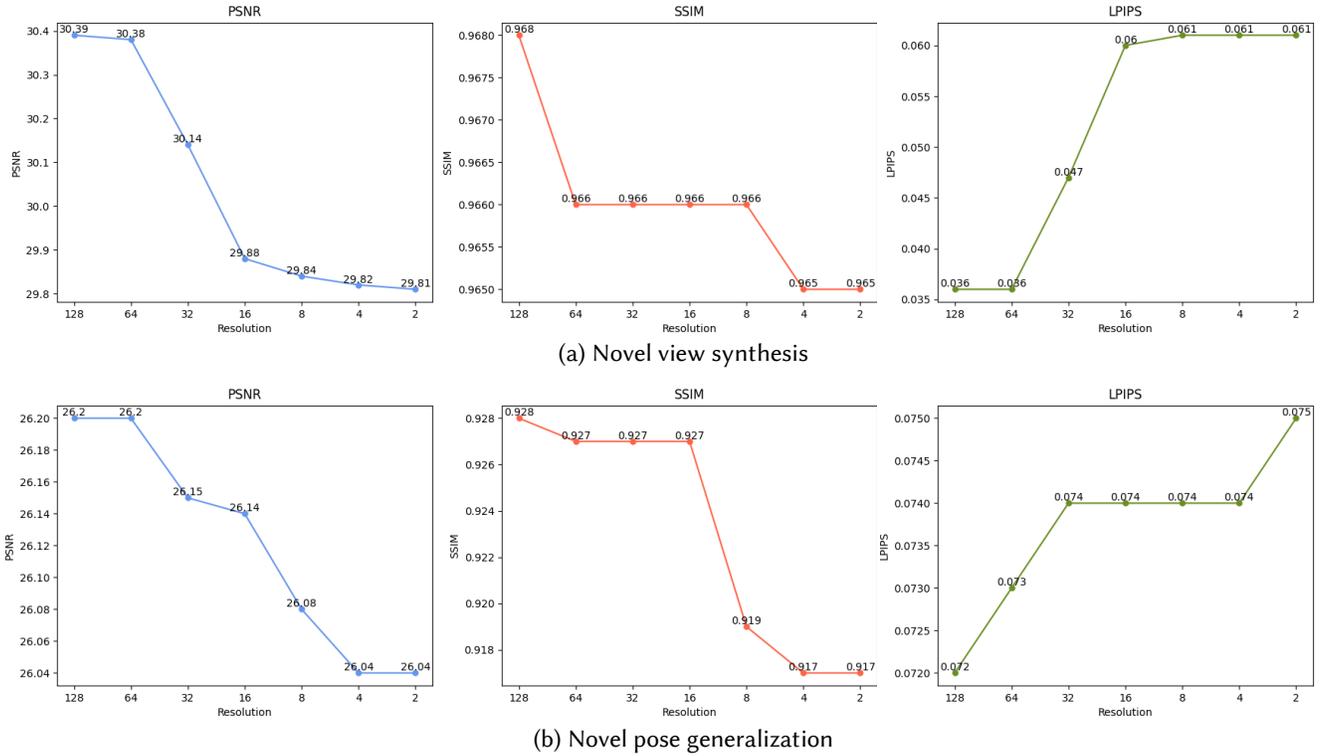}
  \caption{Impacts of NTS at different resolutions.}
\label{fig:ab_res}
\end{figure*}

\noindent\textbf{NTS at Different Resolutions.} As shown in Figure \ref{fig:conv_texture}, we generate an NTS at a resolution of $64 \times 64$. Choosing the resolution of NTS provides a trade-off between quality and memory. 
We analyze the impacts of resolution in Figure \ref{fig:ab_res}, where we report test quality vs. resolution for the dataset of CMU-p2 on PSNR, SSIM and LPIPS metrics. 
Restricted to memory limitations, NTS has a maximum resolution of 128. 
It can be observed that the larger resolution of NTS, the better the model performed on novel view synthesis and the novel pose generalization tasks.
In this analysis, we found $64 \times 64$ to be a favorable optimum in our applications, where NTS at $128 \times 128$ resolution is not much better than at $64 \times 64$ resolution but costs more memory and time, so we choose $64 \times 64$ resolution in all other experiments and recommend it as the default to practitioners.

\subsection{View-dependent Color}
\noindent\textbf{Ray Direction Noise.}
To model view-dependent RGB color of human performances, we apply positional encoding $\gamma(\cdot)$~\cite{rahaman2019spectral} to the viewing direction, and pass the encoded viewing direction, UV map and the sampled texture embedding into the color network $M_c$ to decode the view-dependent color $\hat{\mathcal{C}}_{k}(\mathbf{r})$ of camera ray $\mathbf{r}$. 

Since the UV map is generated in a learning-based fashion rather than using direct sampling locations, the color network tends to overfit training viewing directions directly sampled during training. To improve the generalisability of the color network, we apply a sub-pixel noise to the ray direction. Here, instead of shooting in the pixel centers, a noise $\psi$ is used as follows:
\begin{equation}
x_{i}=\mathbf{o}+t_{i}(\mathbf{d}+\psi),
\end{equation}
where noise $\psi$ can be interpreted as a locality condition, i.e., in similar view conditions, RGB color should not be too different. It allows the model to learn smoother transitions between different views.

\begin{figure*}[t]
\centering
  \includegraphics[width=1\linewidth]{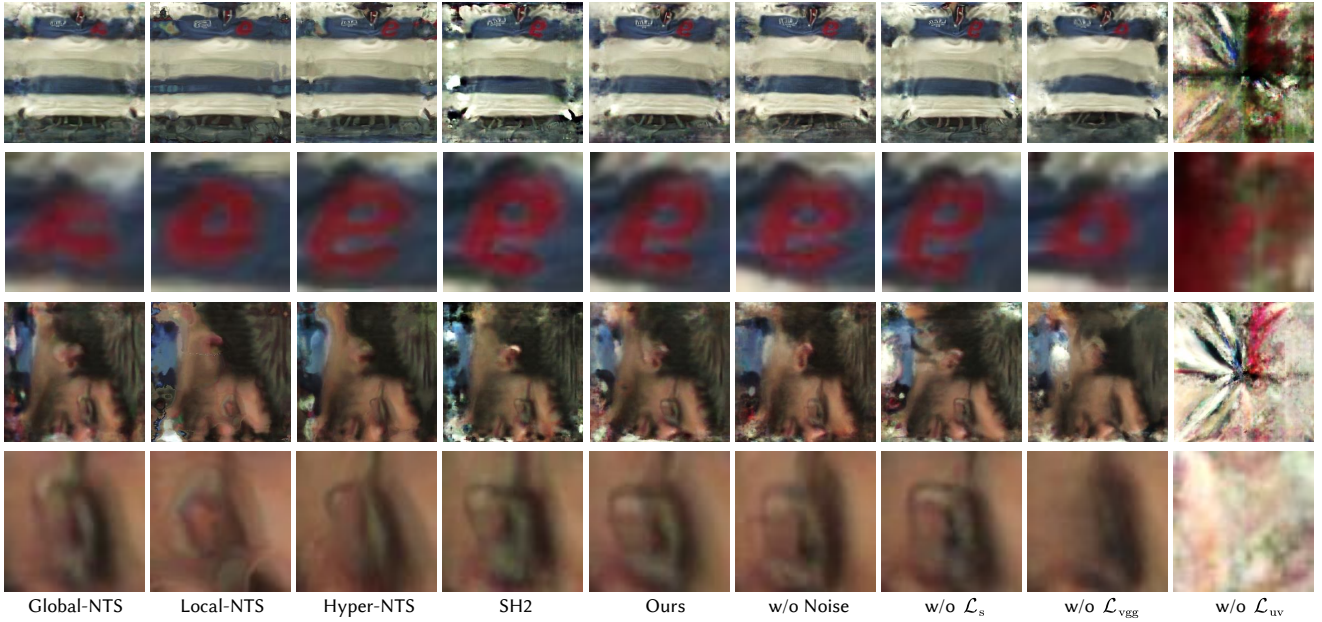}
  \caption{Textures of our model against ablations.}
\label{fig:ablation_texture}
\end{figure*}

The ablation of ours w/o noise is presented in Table \ref{tab:view} and demonstrates the effectiveness of the proposed ray direction noise. Figure \ref{fig:ablation_rgb} and Figure \ref{fig:ablation_texture} show qualitative results of ours w/o noise tests on the novel views. Obviously, ours w/o noise tends to exhibit artifacts in the rendering and textures.

\begin{table}[t]
\centering
\setlength\tabcolsep{3pt}
\resizebox{\linewidth}{!}{
\begin{tabular}{c||ccc|ccc}
\toprule
\multirow{2}{*}{Type} & \multicolumn{3}{c|}{Novel view synthesis} & \multicolumn{3}{c}{Novel pose generation} \vspace{1.5pt} \\
& \small PSNR$\uparrow$ 
& \small SSIM$\uparrow$ 
& \small LPIPS$\downarrow$ 
& \small PSNR$\uparrow$ 
& \small SSIM$\uparrow$ 
& \small LPIPS$\downarrow$ \\
\midrule
SH1 & 29.41 & \cellcolor{second}{0.966} & 0.052 & 26.00 & 0.926 & 0.078 \\
SH2 & 30.02 & \cellcolor{second}{0.966} & 0.051 & 26.15 & \cellcolor{second}{0.927} & 0.076 \\
SH3 & 29.43 & 0.965 & 0.051 & 26.14 & 0.926 & 0.075 \\
SH4 & 27.83 & 0.962 & 0.054 & 26.10 & 0.926 & 0.076 \\
w/o noise & 28.56 & 0.962 & 0.058 & 25.90 & 0.924 & 0.078 \\
Ours & \cellcolor{second}{30.38} & \cellcolor{second}{0.966} & \cellcolor{second}{0.036} & \cellcolor{second}{26.20} & \cellcolor{second}{0.927} & \cellcolor{second}{0.073} \\
\bottomrule 
\end{tabular}
}
\caption{Comparison of different methods to model the view-dependent RGB color.}
\label{tab:view}
\end{table}

\noindent\textbf{Spherical Harmonic Functions.}
Another way to reconstruct the view-dependent color of human performance is using the spherical harmonic (SH) functions. We pass the encoded UV map with the sampled texture embedding into the color network $M_c$ to decode spherical harmonic coefficients $\eta$ for each color channel:
\begin{align}
(\hat{\eta}^{0}_{k}(\mathbf{r}), \hat{\eta}^{1}_{k}(\mathbf{r}), \cdots, & \hat{\eta}^{n}_{k}(\mathbf{r}))= \\
\nonumber
& {M}_{c}\!\left(\gamma(\hat{\mathcal{U}}_{k}(\mathbf{r}),\hat{\mathcal{V}}_{k}(\mathbf{r})),\mathbf{e}_{k}(\mathbf{r}),\mathbf{k}\right),
\end{align}
where spherical harmonics $(\eta^{0}, \eta^{1}, \cdots, \eta^{n})$ form an orthogonal basis for functions defined over the sphere, with zero degree harmonics $\eta^{0}$ encoding diffuse color and higher degree harmonics encoding specular effects. The view-dependent color $\hat{\mathcal{C}}_{k}(\mathbf{r})$ of camera ray $\mathbf{r}$ can be determined by querying the specular spherical functions $SH$ at desired viewing direction $\mathbf{d}$:
\begin{equation}
\hat{\mathcal{C}}_{k}(\mathbf{r})=S\left(\frac{\hat{\eta}^{0}_{k}(\mathbf{r})}{2}\sqrt{\frac{1}{\pi}} + \sum_{m=1}^{n} SH^{m}(\hat{\eta}^{m}_{k}(\mathbf{r}), \mathbf{d})\right),
\end{equation}
where $S$ is the sigmoid function for normalizing the colors.

A higher degree of harmonics results in a higher capability to model high-frequency color but is more prone to overfit the training viewing direction. 

The ablation of different harmonics degrees is presented in Table \ref{tab:view}, which demonstrates that the harmonics degree of $2$ model achieves the best performance among all the SH models but still cannot reach the performance of ours. Figure \ref{fig:ablation_rgb} illustrates the qualitative results of SH2 tests on the novel views, which shows that SH2 suffers from global color shifts (head of the performer) and artifacts (the belt on pants), while ours does not.

\subsection{Video Frames and Input Views}
To analyze the impacts of the number of camera views and video length, we show the results of our models trained with different numbers of camera views and video frames in Table \ref{tab:view_frames}. We conduct the experiments on performer 313 of the ZJU dataset. All the results are evaluated on the rest two views of the first 60-frame video. The results show that although the number of training views improves the performance on novel view synthesis, sparse four views are good enough for our model to reconstruct dynamic human performances. 
In addition, the ablation study of frame numbers indicates that training on too many frames may decrease the performance as the network cannot fit such a long video, which is also mentioned in NeuralBody~\cite{peng2021neural}.

\section{Additional Results} 
\subsection{Novel View Synthesis}
\begin{figure*}[t]
\centering
  \includegraphics[width=1\linewidth]{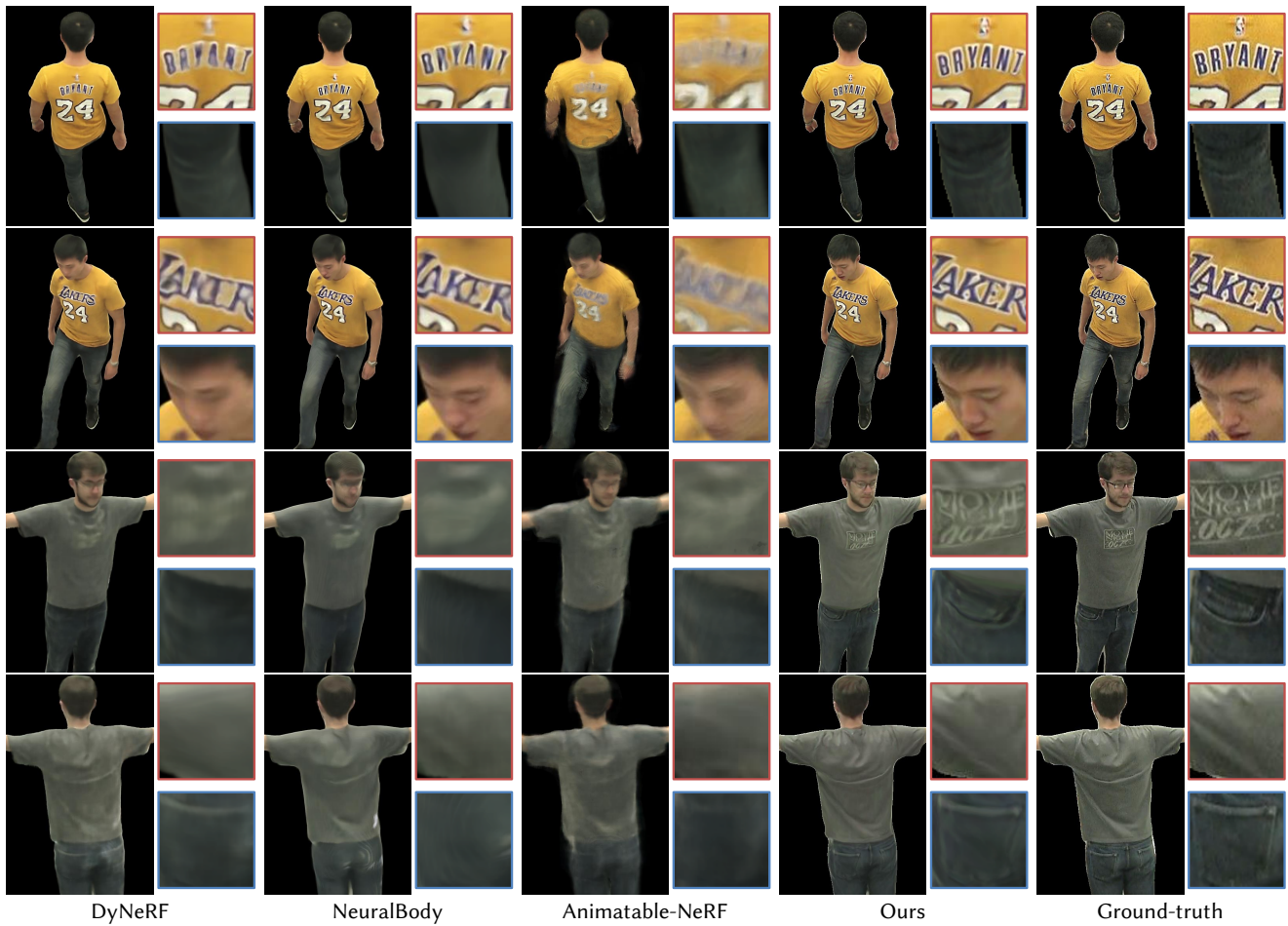}
  \caption{Comparisons on test-set views for performers from CMU Panoptic dataset with 960$\times$540 images. Our model generates photo-realistic appearance images even with rich textures, particularly letters on the performers' clothes. By contrast, baselines give blurry results while missing a lot of high-frequency details. Here, we present results on two different test views at the same time for each performer.}
\label{fig:supp_nvcmu}
\end{figure*}

\begin{figure*}[t]
\centering
  \includegraphics[width=1\linewidth]{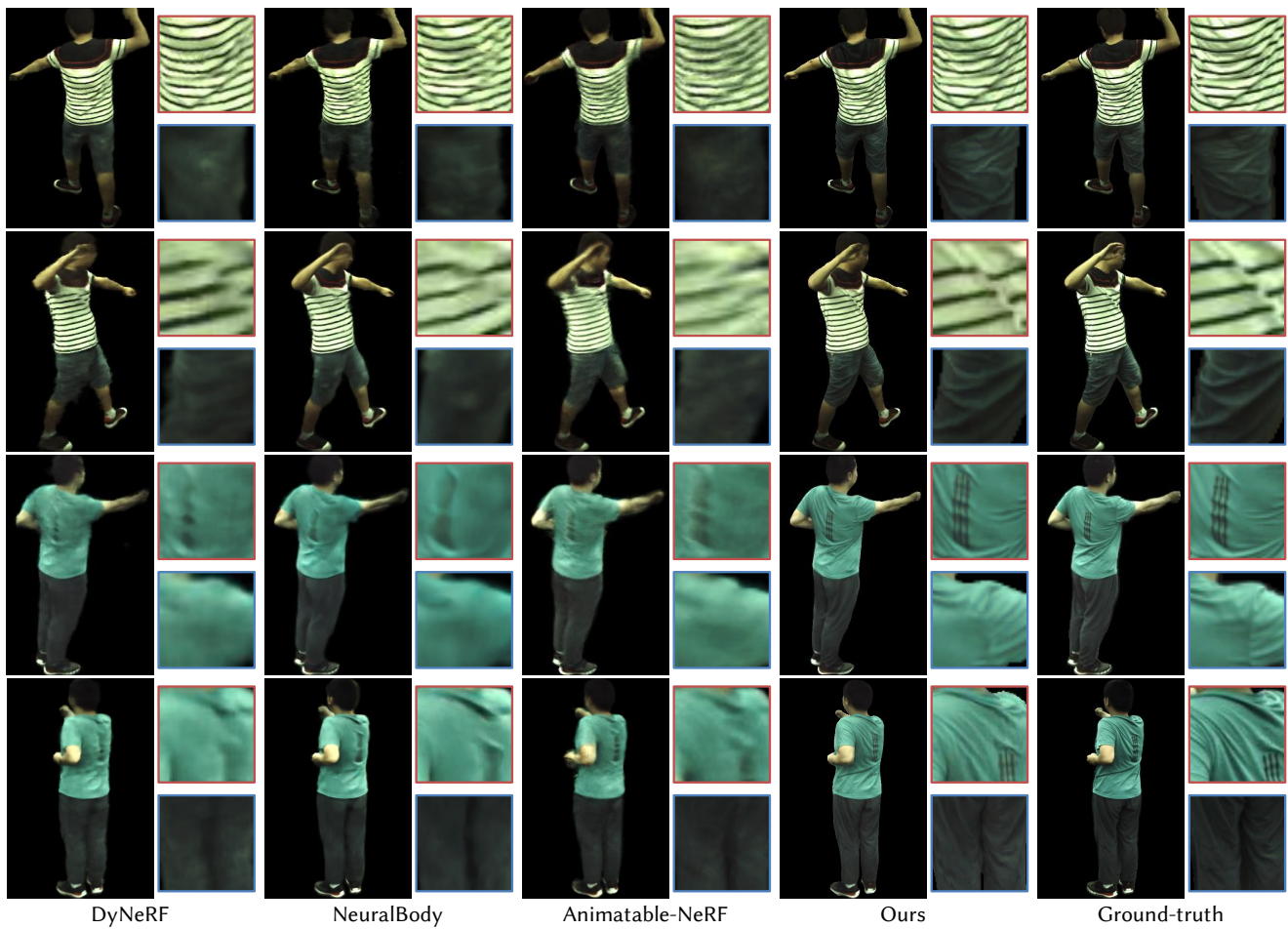}
  \caption{Comparisons on test-set views for performers from the ZJU Mocap dataset. Our model obviously performs well in details (e.g., stripes on T-shirts and wrinkles in clothes) and sharpness than all other baselines, which benefits from our proposed Spatial NTS that encodes high-frequency appearance information. Other methods give plausible but blurry and rough synthesized images. Here, we present results on two different test views at the same time for each performer.}
\label{fig:supp_nvzju}
\end{figure*}

\begin{figure*}[t]
\centering
  \includegraphics[width=1\linewidth]{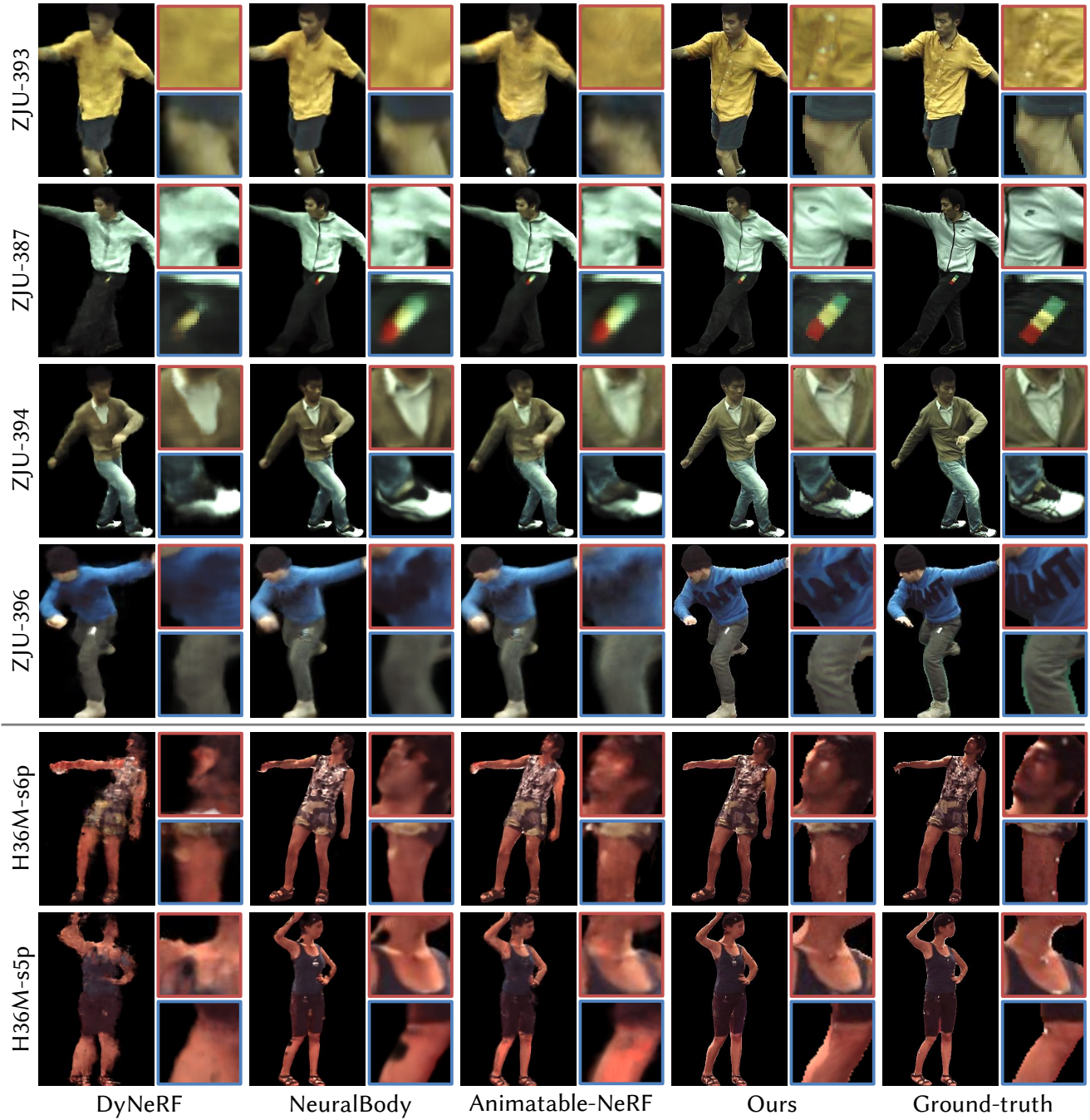}
  \caption{Comparisons on test-set views from ZJU Mocap dataset with four training views and the most challenging H36M dataset with only three available views for training. Our model generates high-definition results even with rich textures and challenging motions. DyNeRF fails to render plausible results with sparse training views because taking time-varying latent codes as the conditions are hard to reuse information among frames.}
\label{fig:supp_nvh36m}
\end{figure*}

For comparison, we synthesize images of training poses in hold-out test-set views. More qualitative results of novel view synthesis are shown in Figure \ref{fig:supp_nvcmu}, Figure \ref{fig:supp_nvzju} and Figure \ref{fig:supp_nvh36m}. Our method produces photo-realistic images with sharp details, particularly letters on clothes (in Figure \ref{fig:supp_nvcmu}), stripes on T-shirts, and wrinkles in clothes (in Figure \ref{fig:supp_nvzju}), which benefits from our proposed Spatial NTS that encodes high-frequency appearance information.

\begin{table}[tb]
\centering
\setlength\tabcolsep{3pt}
\resizebox{\linewidth}{!}{
\begin{tabular}{c||ccc|ccc}
\toprule
\multirow{2}{*}{Task} & \multicolumn{3}{c|}{4 Views} & \multicolumn{3}{c}{20 Views} \vspace{1.5pt} \\ 
& \small PSNR$\uparrow$ 
& \small SSIM$\uparrow$ 
& \small LPIPS$\downarrow$ 
& \small PSNR$\uparrow$ 
& \small SSIM$\uparrow$ 
& \small LPIPS$\downarrow$ \\
\midrule
60 Frames & 29.56 & 0.967 & 0.045 & 31.12 & 0.975 & 0.038 \\
300 Frames & 29.23 & 0.963 & 0.048 & 29.90 & 0.968 & 0.046 \\
600 Frames & 29.37 & 0.964 & 0.049 & 29.50 & 0.966 & 0.048 \\
1200 Frames & 28.96 & 0.961 & 0.052 & 29.26 & 0.963 & 0.053 \\
\bottomrule 
\end{tabular}
}
\caption{Ablation study on the number of training frames and views.}
\label{tab:view_frames}
\end{table}

Figure \ref{fig:supp_nvh36m} shows the results of comparisons on ZJU Mocap and H36M dataset, which are trained on sparse-views video sequences. Here, we use four training views on ZJU Mocap dataset and three for the most challenging H36M dataset. Our model obviously performs much better in details and sharpness than all other baselines. Furthermore, DyNeRF fails to render plausible results with sparse training views because taking time-varying latent codes as the conditions are hard to reuse information among frames.

\subsection{Novel View Synthesis of Dynamic Humans}
\begin{figure*}[t]
\centering
  \includegraphics[width=1\linewidth]{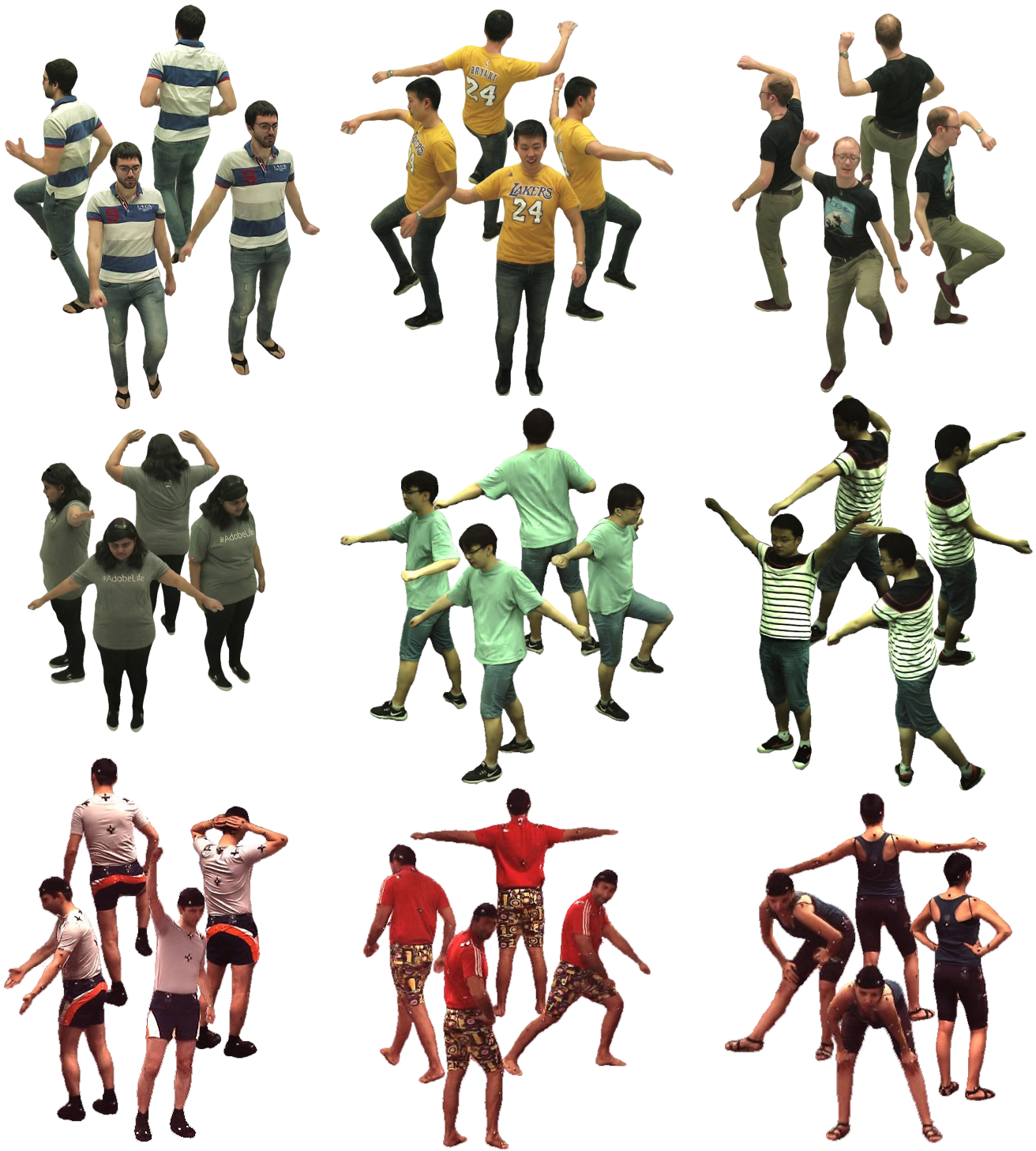}
  \caption{The rendering of our method on different sequences. Our model can handle a dynamic human with rich textures and challenging motions preserving sharp image details like letters and wrinkles, which benefits from our proposed Spatial NTS that encodes high-frequency appearance information 
  while keeping inter-view and inter-frame consistency, which benefits from our proposed \textit{UV Volumes}. Note that the last row is the result of our model on the H36M dataset, which demonstrates that our model can still recover high-fidelity free-view videos under sparse training views.}
\label{fig:dynamic360}
\end{figure*}

We present more results on novel view synthesis of dynamic humans in Figure \ref{fig:dynamic360}. As presented, our model can handle a dynamic human with rich textures and challenging motions, and preserve sharp image details like letters and wrinkles, while keeping inter-view and inter-frame consistency. Note that the last row is the result of our model on the H36M dataset, demonstrating that our model can still recover high-fidelity free-view videos under sparse training views.

\begin{figure*}[t]
\centering
  \includegraphics[width=1\linewidth]{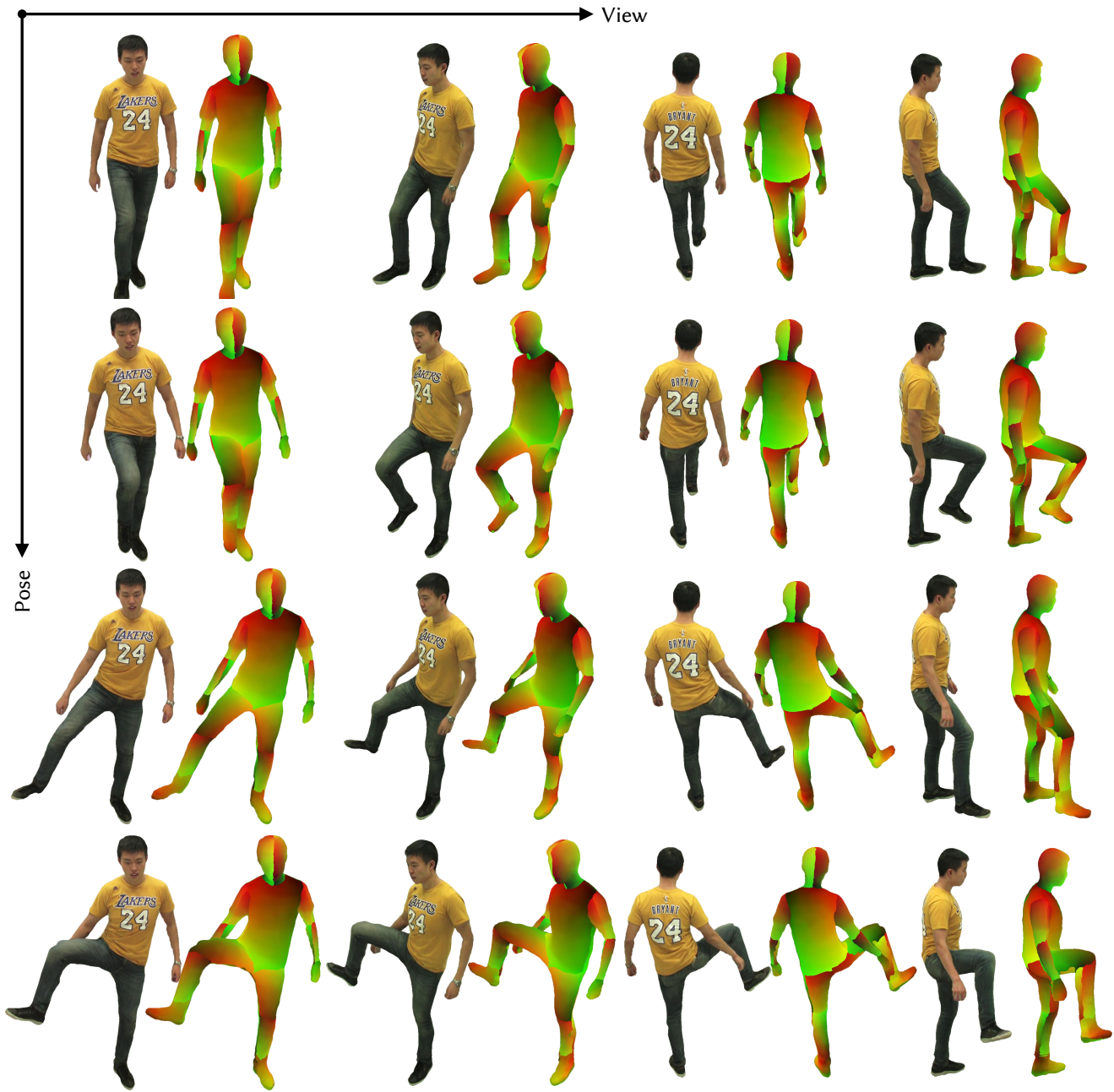}
  \caption{Novel view synthesis of the dynamic human. Our model can synthesize photo-realistic view-consistent RGB images condition on view-consistent UV images rendered by UV volumes. Here, The horizontal axis shows the change in novel views and, the vertical axis shows the change in human poses. All results are rendered from novel views in the training pose sequence.}
\label{fig:iuv_view_pose_change}
\end{figure*}

In addition, we show the intermediate UV images and final RGB images rendered by our model varying with views and human poses in Figure \ref{fig:iuv_view_pose_change}, which demonstrates that our model can synthesize photo-realistic view-consistent RGB images that condition on view-consistent UV images rendered by UV volumes.

\subsection{Novel Pose Generalization}
\begin{figure*}[tb]
\centering
  \includegraphics[width=0.85\linewidth]{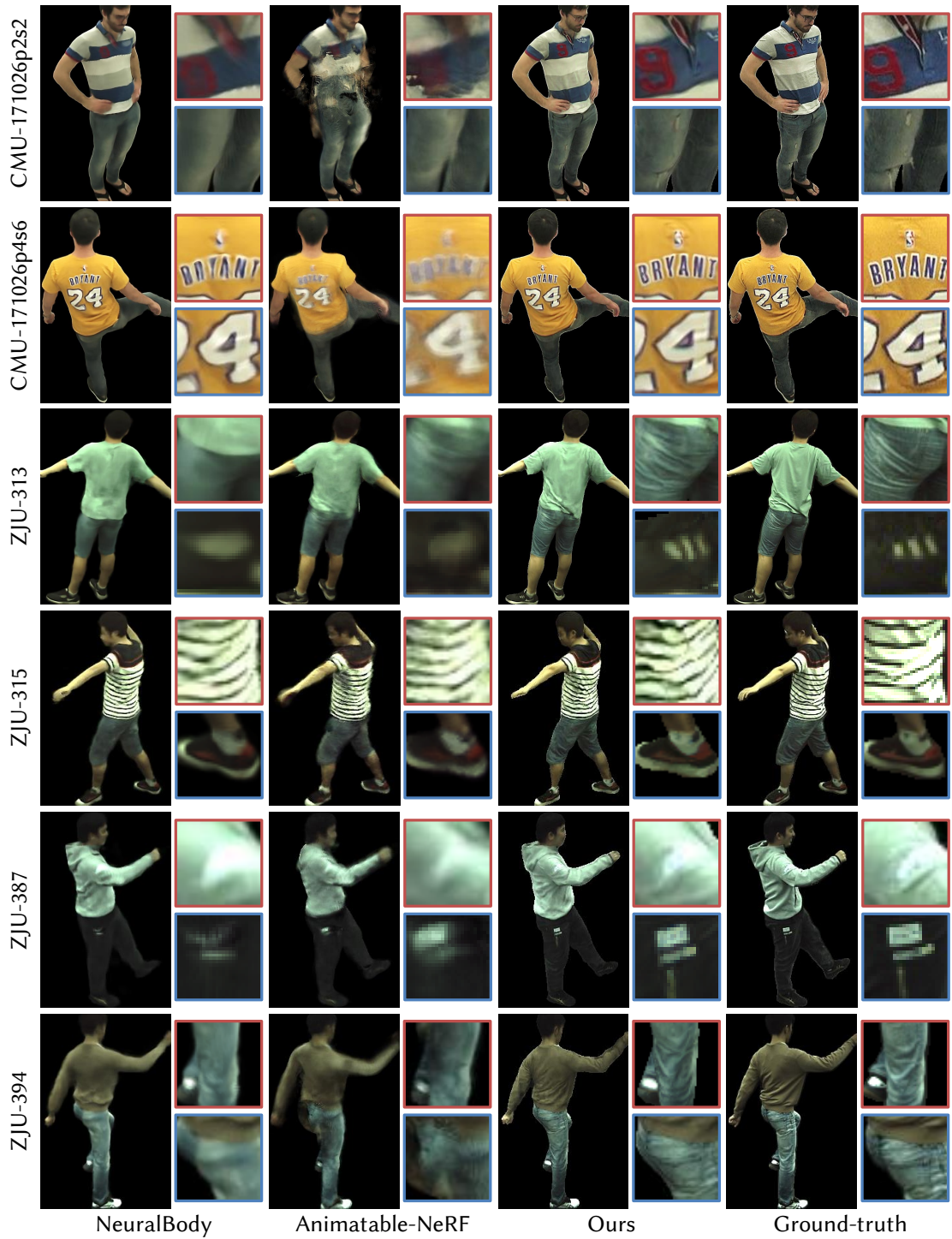}
  \caption{Comparisons on test-set poses for performers from CMU Panoptic and ZJU Mocap dataset.}
\label{fig:novel_pose_add1}
\end{figure*}

\begin{figure*}[tb]
\centering
  \includegraphics[width=0.85\linewidth]{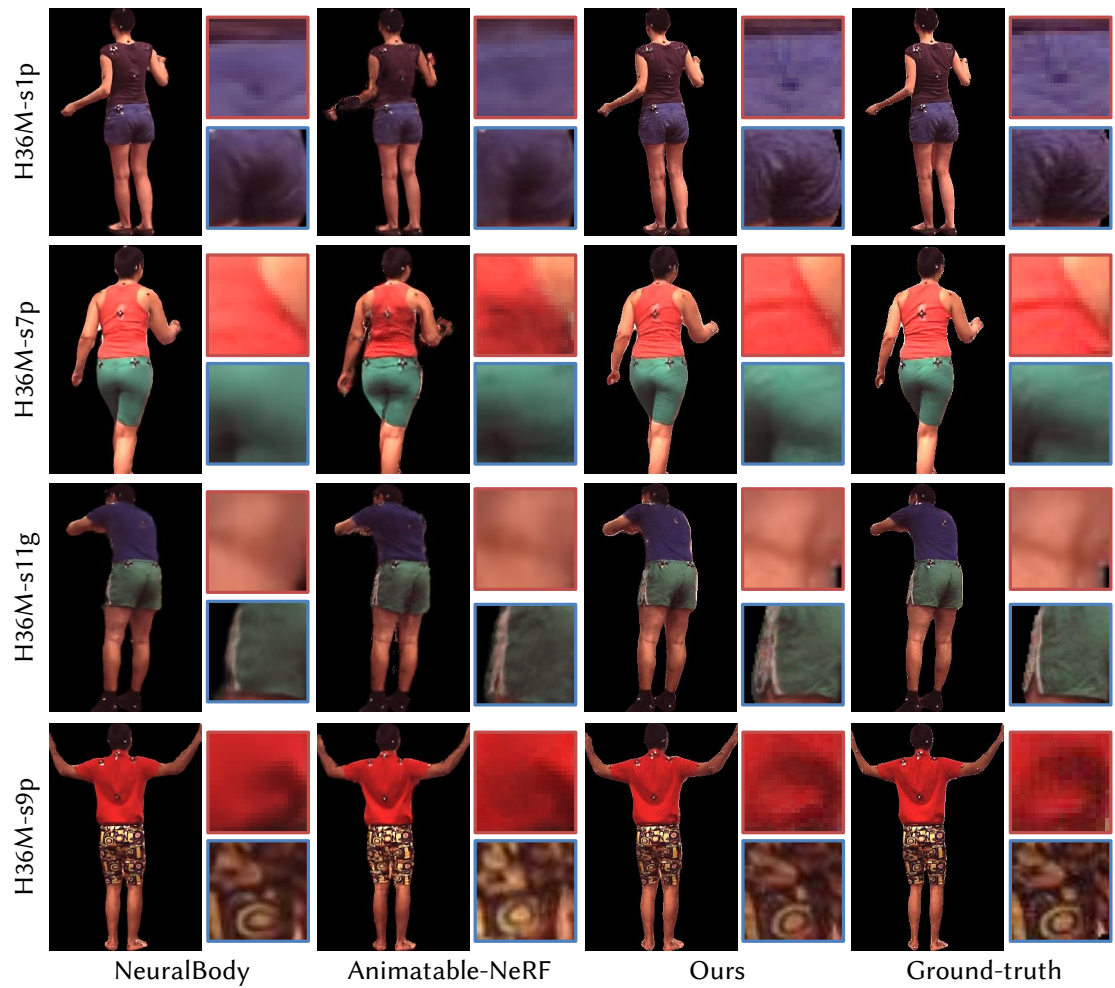}
  \caption{Comparisons on test-set poses for performers from the most challenging H36M dataset.
  }
\label{fig:novel_pose_add2}
\end{figure*}

More qualitative results of novel pose generalization are shown in Figure \ref{fig:novel_pose_add1} and Figure \ref{fig:novel_pose_add2}, where the latter are the results of comparisons on H36M dataset where only three cameras are available for training.

\subsection{Reshaping}
\begin{figure*}[tb]
\centering
  \includegraphics[width=1\linewidth]{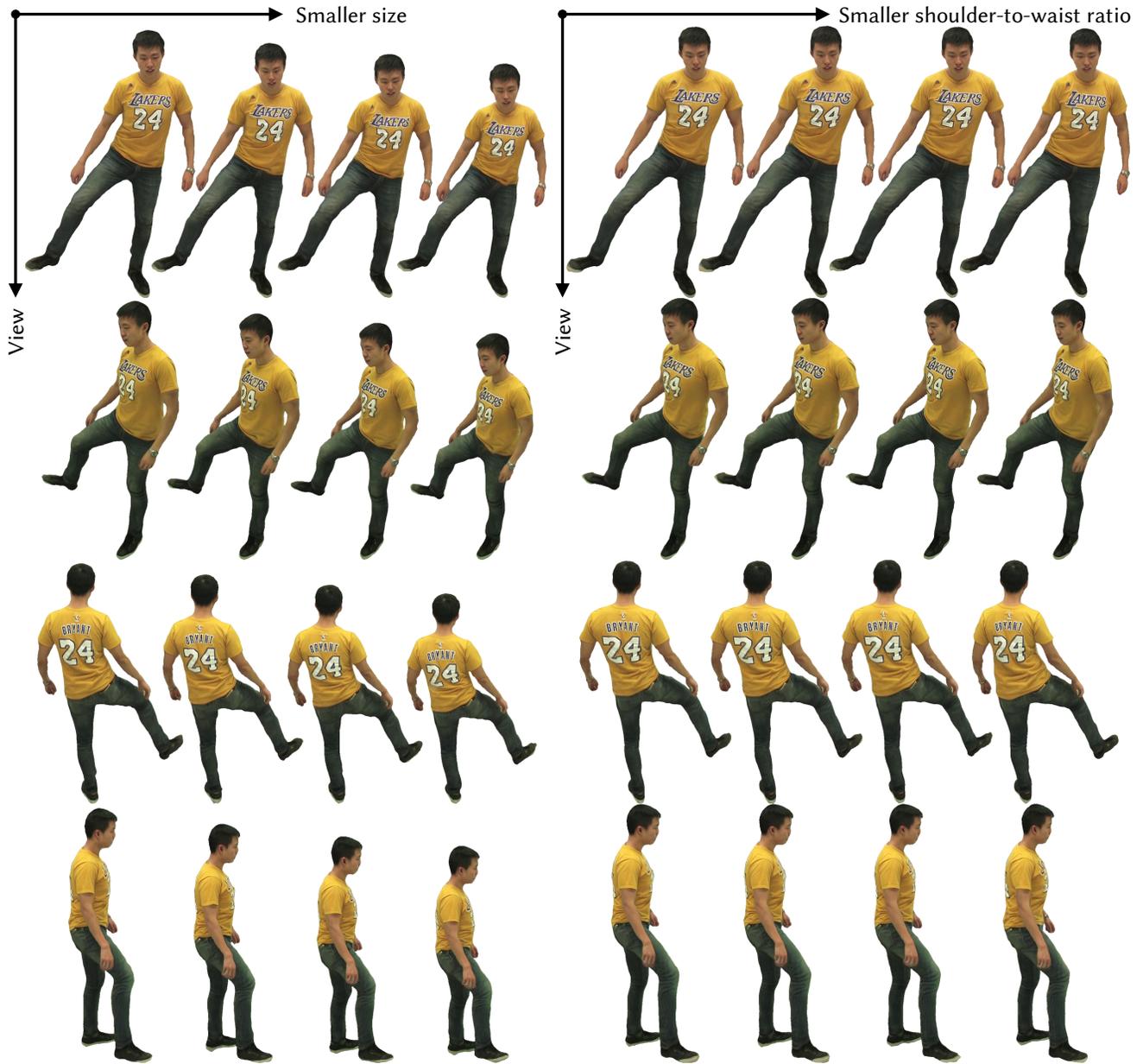}
  \caption{Novel view synthesis results of reshaping. By changing the SMPL parameters, we can conveniently deform the human performer. We present the performer whose size is getting smaller and the shoulder-to-waist ratio is getting smaller from left to right. With the help of view-consistent UV coordinates encoded by UV volumes, our model still renders view-consistent images with challenging shape parameters. Here, The horizontal axis shows shape changes and the vertical axis shows view changes. All results are rendered from novel views.}
\label{fig:reshape_view}
\end{figure*}

By changing the SMPL parameters, we can conveniently deform the human performer. We present the performer whose size is getting smaller and the shoulder-to-waist ratio is getting smaller from left to right in Figure \ref{fig:reshape_view}. With the help of view-consistent UV coordinates generated by UV volumes, our model still renders view-consistent images with challenging shape parameters. These rendered images maintain high appearance consistency across changing shapes thanks to the neural texture stacks.

\subsection{Visualization of NTS}
\begin{figure*}[tb]
\centering
  \includegraphics[width=1\linewidth]{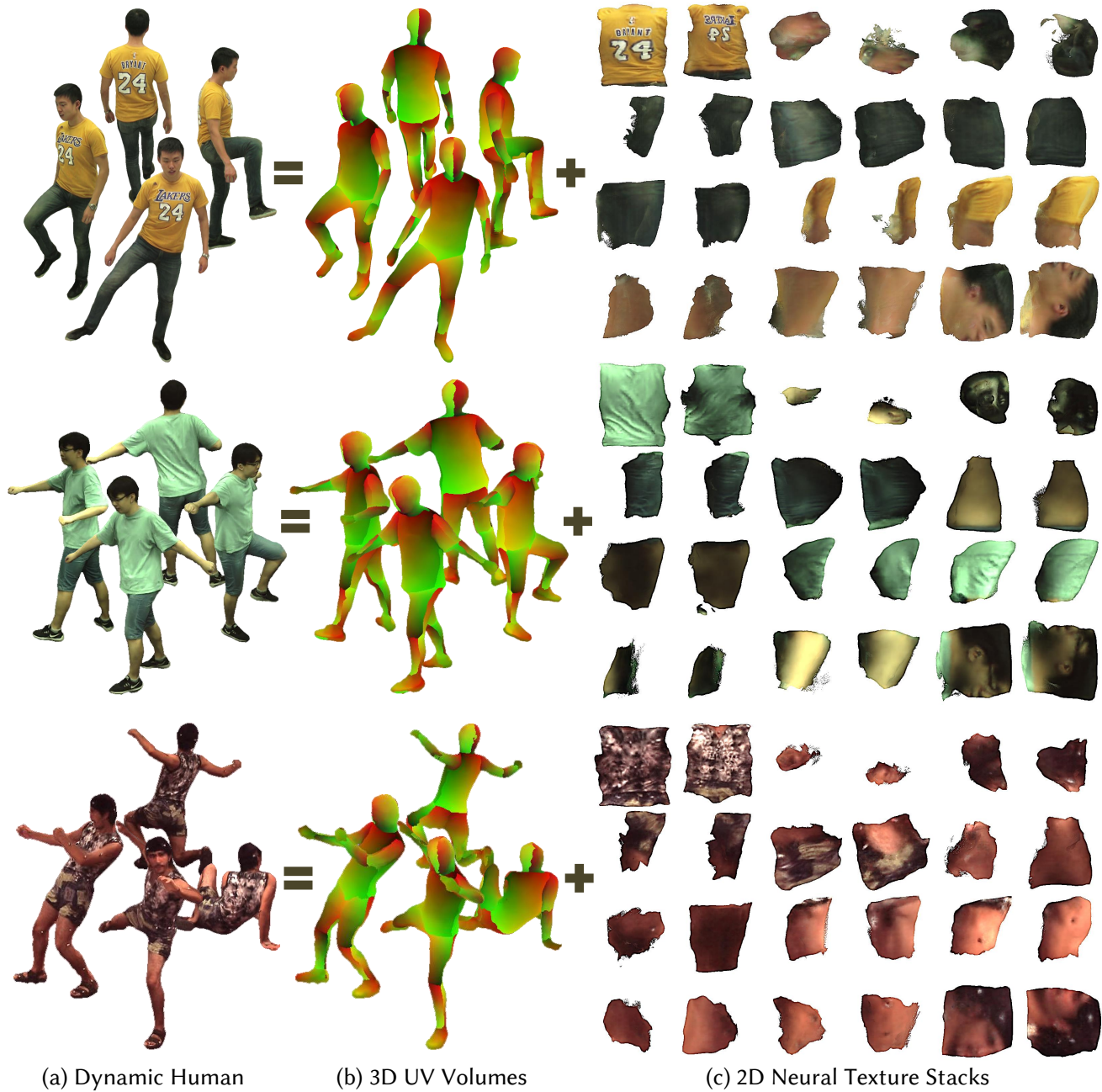}
  \caption{We decompose (a) the dynamic human into (b) 3D UV volumes and (c) 2D neural texture stacks. The disentanglement of appearance from geometry enables us to achieve real-time rendering of free-view human performance. We show performers and their UV avatars with four different poses at four different viewing directions from CMU Panoptic, ZJU-Mocap and H36M datasets. Their neural texture stacks that preserve human appearance with high-frequency details under one of these poses are visualized in the last column. Our method takes smooth UV coordinates to sample neural texture stacks for corresponding RGB value.}
\label{fig:dy360_texture}
\end{figure*}

\begin{figure*}[tb]
\centering
  \includegraphics[width=1\linewidth]{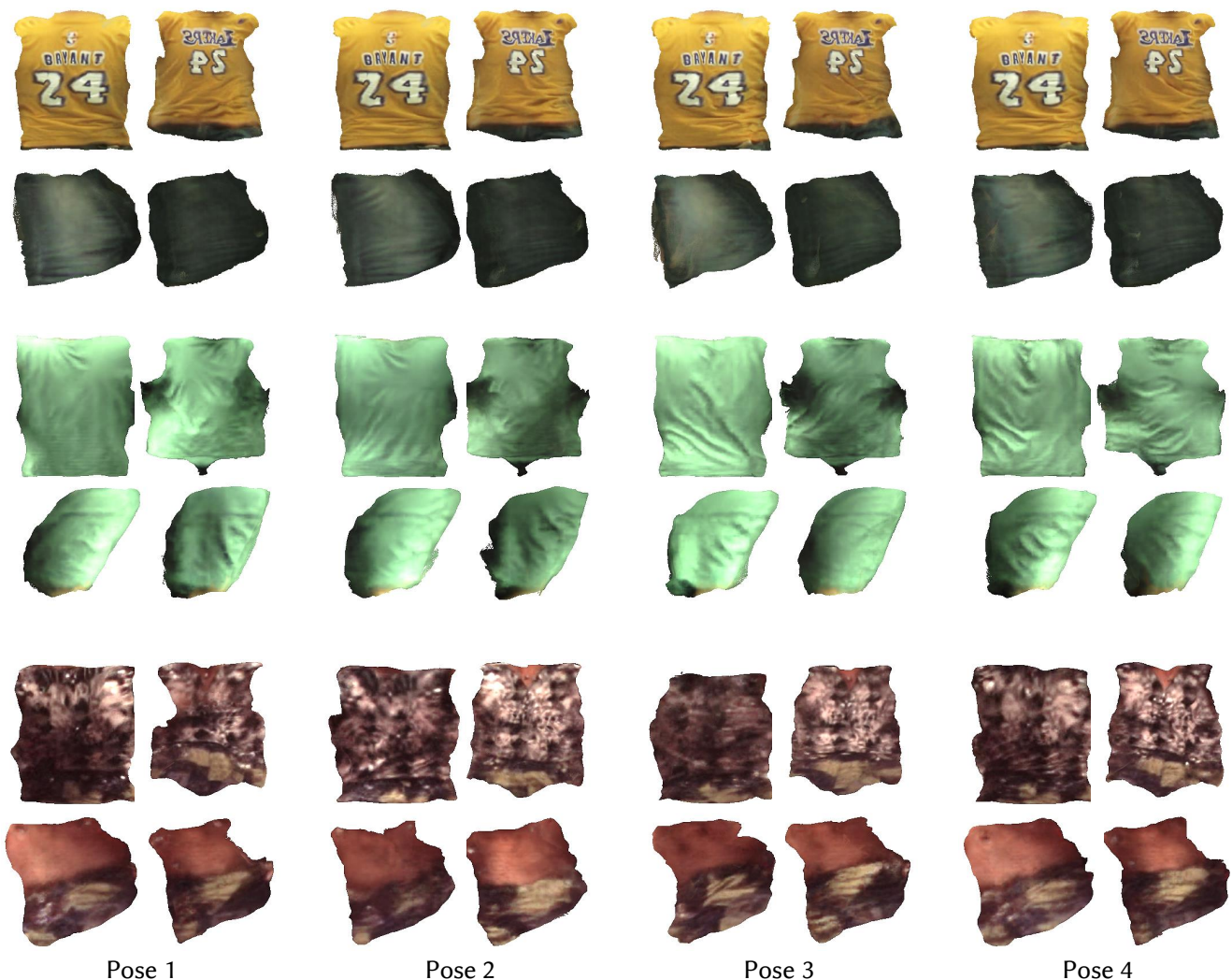}
  \caption{Visualization of neural texture stacks under different poses. Details like the folds of clothing vary from motion to motion, as does the topology. Therefore, we propose pose-driven neural texture stacks to describe textures at different times, which enables us to handle dynamic 3D reconstruction tasks and to generalize our model to unseen poses.}
\label{fig:pose_dependent_texture}
\end{figure*}

\vspace{-9pt}
In contrast to ~\cite{peng2021neural} learning a radiance filed in the 3D volumes, we decompose a 3D dynamic human into 3D UV volumes and 2D neural texture stacks, as illustrated in Figure \ref{fig:dy360_texture}. The disentanglement of appearance from geometry enables us to achieve real-time rendering of free-view human performance. We learn a view-consistent UV field to transfer neural texture embeddings to colors, which guarantees view-consistent human performance. Details like the folds of clothing vary from motion to motion, as does the topology, so we require a dynamic texture representation. Referring to Figure \ref{fig:pose_dependent_texture}, we visualize the pose-driven neural texture stacks to describe appearance at different times, which enables us to handle dynamic 3D reconstruction tasks and to generalize our model to unseen poses. It is obvious that our learned NTS preserve rich textures and high-frequency details varying from different poses.
\vspace{10pt}

\subsection{Retexturing}
\begin{figure*}[tb]
\centering
  \includegraphics[width=1\linewidth]{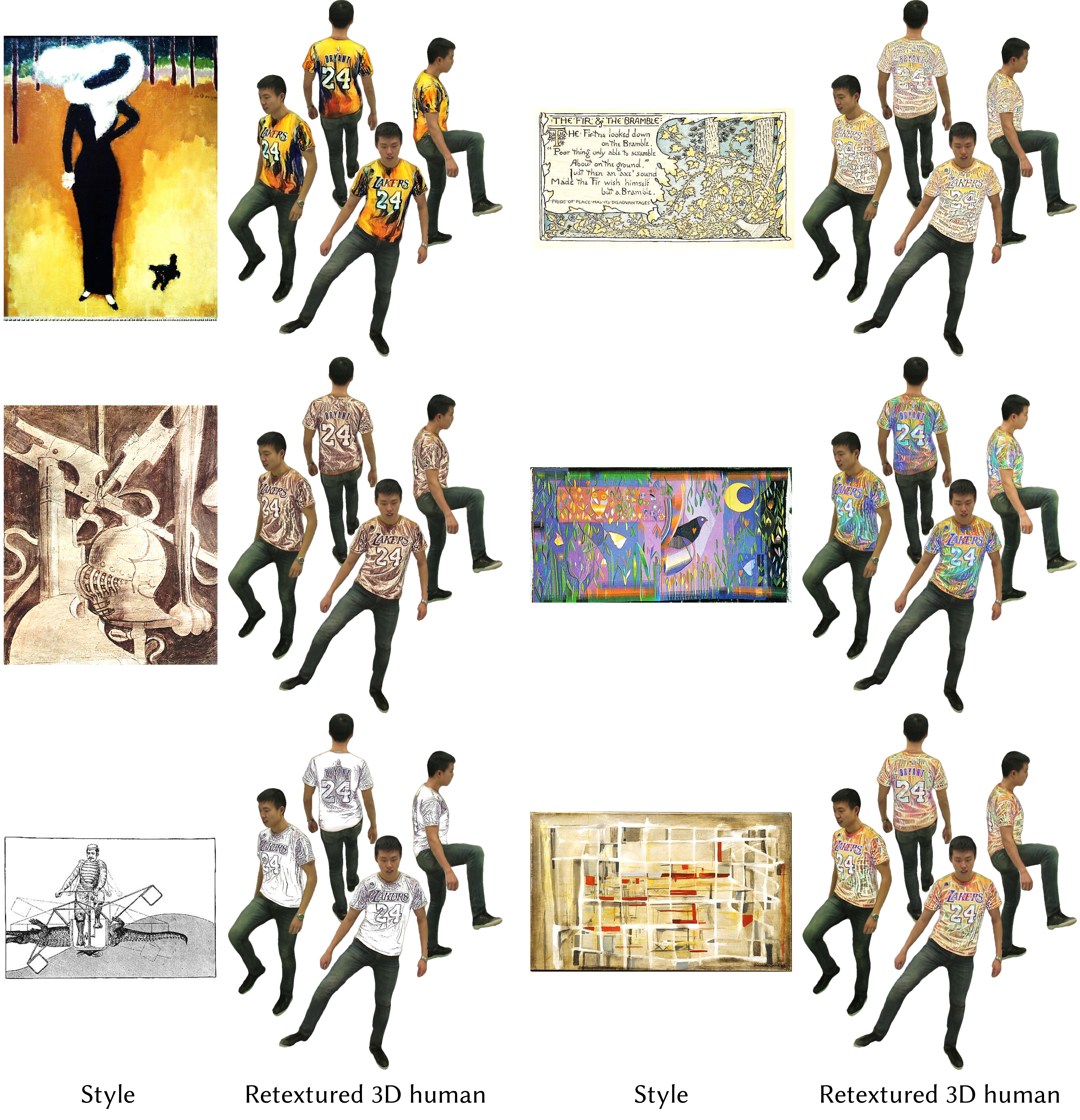}
  \caption{Our model supports rendering a stylized dynamic human with arbitrary artistic painting, which can be applied in controllable 3D style transfer with multi-view consistency.}
\label{fig:more_transfer}
\end{figure*}

\begin{figure*}[tb]
\centering
  \includegraphics[width=1\linewidth]{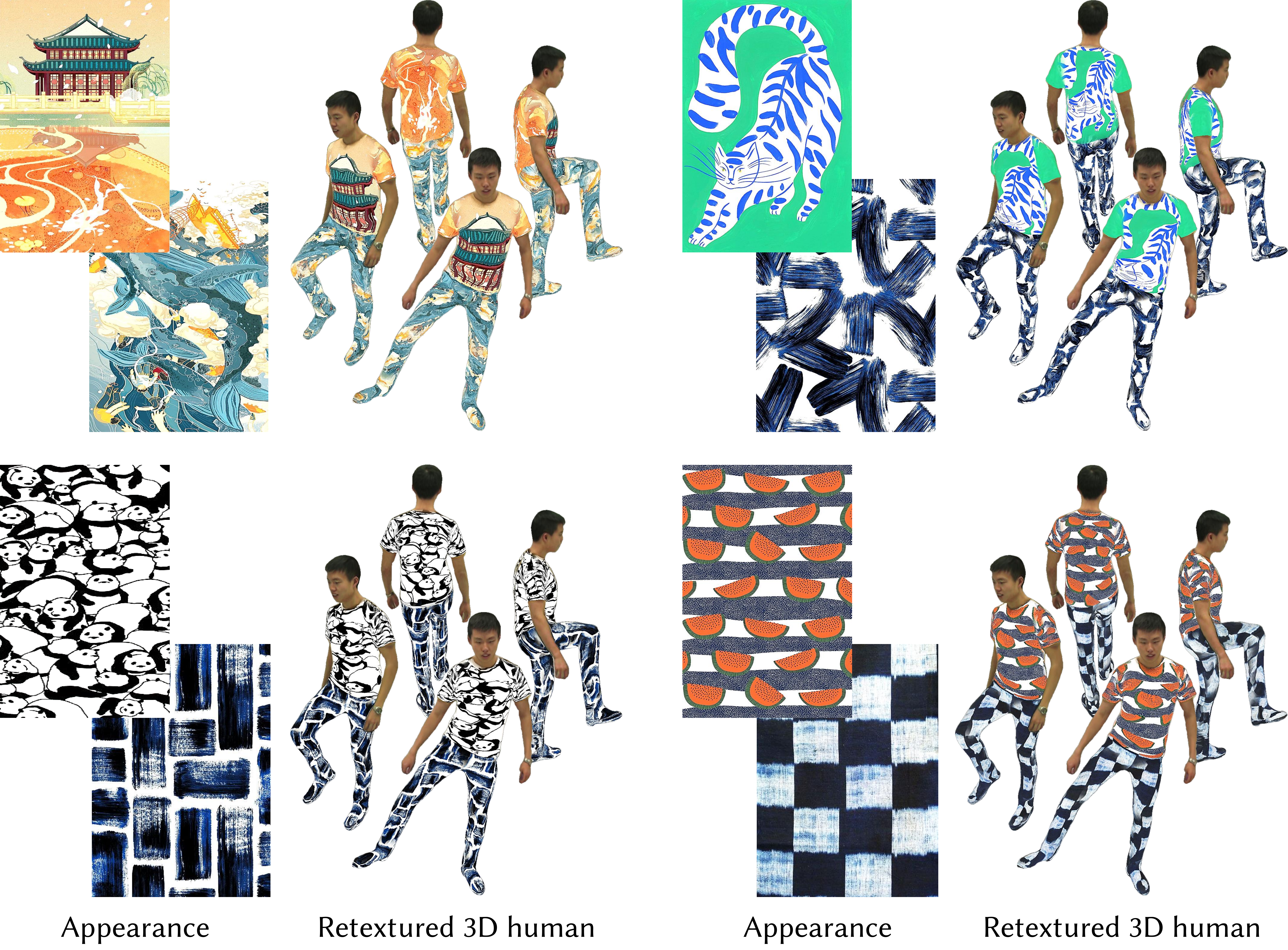}
  \caption{Our model allows us to generate free-view human performance with a user-provided cloth texture image, which enables some interesting applications such as real-time 3D virtual try-on. We collect these appearance images from the Internet. 
  }
\label{fig:more_new_texture}
\end{figure*}

\begin{figure*}[tb]
\centering
  \includegraphics[width=1\linewidth]{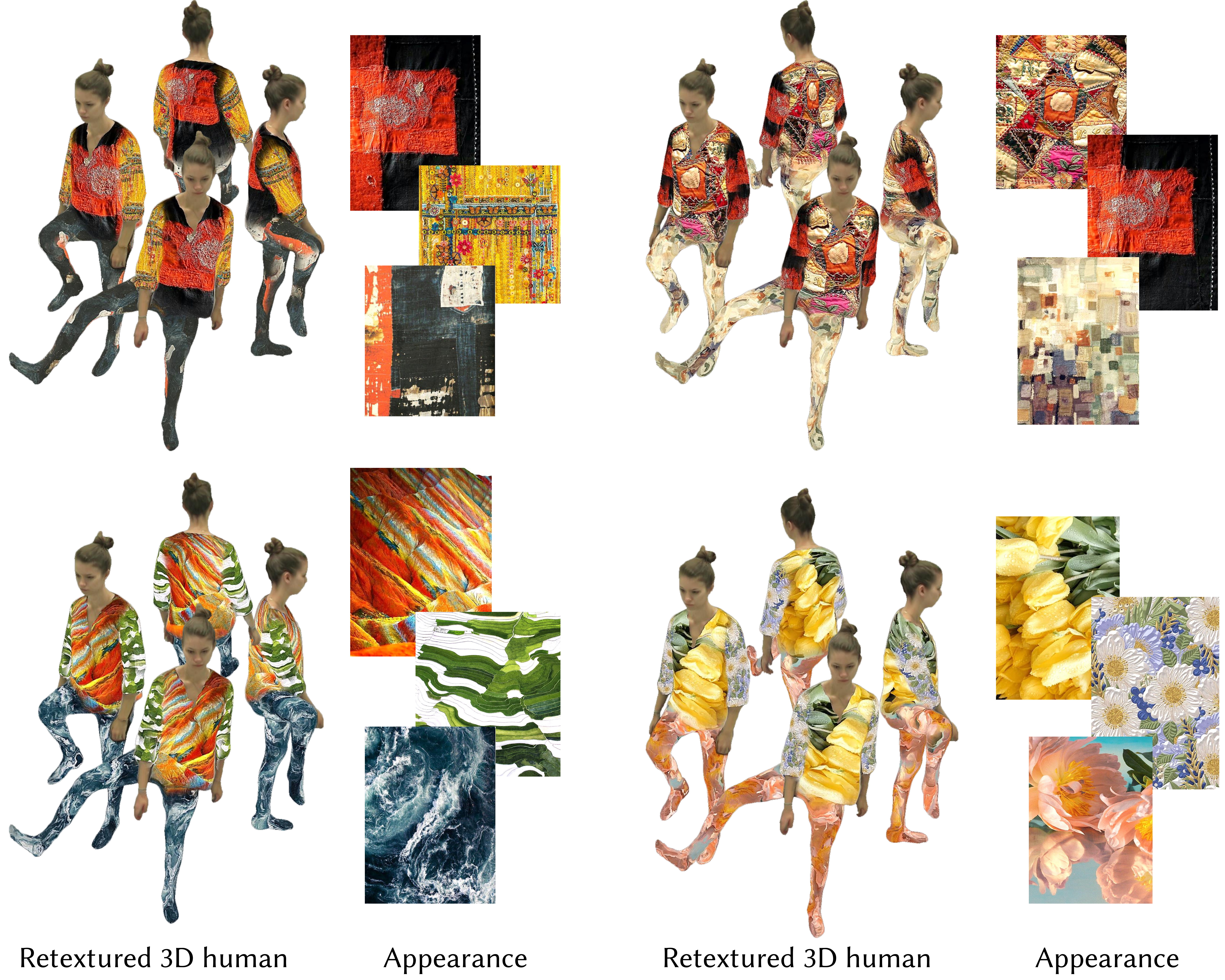}
  \caption{Our model allows us to generate free-view human performance with a user-provided cloth texture image, which enables some interesting applications such as real-time 3D virtual try-on. We collect these appearance images from the Internet. 
  }
\label{fig:more_new_texture2}
\end{figure*}

With the learned dense correspondence of 3D UV volumes and 2D neural texture stacks, we can edit performers' 3D cloth with user-provided 2D textures. As shown in Figure \ref{fig:more_transfer}, given arbitrary artistic paintings, we can produce cool stylized dynamic humans leveraging stylizations transferred from the original texture stacks by the network ~\cite{ghiasi2017exploring}. Visually inspected, the new texture is well painted onto the performer's T-shirt under different poses at different viewing directions. Besides, we perform some interesting applications of our model in Figure \ref{fig:more_new_texture} and Figure \ref{fig:more_new_texture2}, which include a 3D virtual try-on implemented by replacing original texture stacks with a user-provided appearance. The visualization results demonstrate that our model can conveniently edit textures preserving the rich appearance and various styles, which benefits from our proposed Neural Texture Stacks, and can render retextured human performance with view consistency well using 3D UV volumes.

{\small
\bibliographystyle{ieee_fullname}
\bibliography{egbib}
}

\end{document}